%% file: main.tex
\documentclass{article}

\usepackage{iclr2027_conference,times}

\usepackage[utf8]{inputenc}
\usepackage[T1]{fontenc}
\usepackage[hidelinks]{hyperref}
\usepackage{url}
\usepackage{booktabs}
\usepackage{amsmath,amssymb,amsfonts,bm}
\usepackage{nicefrac}
\usepackage{microtype}
\usepackage{xcolor}
\usepackage{color,soul}
\usepackage{graphicx}
\usepackage{multirow}
\usepackage{flafter}
\usepackage{placeins}
\usepackage{enumitem}

\ificlrfinal
\else
\fi

\newcommand{\bq}{\bm q}
\newcommand{\bp}{\bm p}
\newcommand{\bz}{\bm z}

\newcommand{\cL}{\mathcal L}

\title{Improving Hamiltonian Learning through \\Local Energy Decomposition}

\author{%
    Yaojun Li \quad Yulong Yang \quad Christine Allen-Blanchette\\
    Princeton University\\
    \texttt{\{yl3425, yulong.yang, ca15\}@princeton.edu}\\
}

\iclrfinalcopy
\begin{document}

\maketitle
\fancyhead[L]{Preprint}
\begin{abstract}
    \input{sections/0_abstract}
\end{abstract}

\input{sections/1_introduction}
\input{sections/2_related_work}
\input{sections/3_background}
\input{sections/4_method}
\input{sections/5_experiments}
\input{sections/6_conclusion}
\clearpage
\input{sections/n0_statement}

\bibliographystyle{iclr2027_conference}
\bibliography{mul_hnn_references}
\appendix
\input{sections/n1_appendix}

\input{sections/n2_appendix_ot}

\end{document}

%% file: sections/0_abstract.tex
Hamiltonian neural networks provide a principled framework for learning conservative dynamics. A scalar Hamiltonian is parameterized by a neural network and the vector field is derived using Hamilton's equations. However, preserving Hamiltonian structure does not guarantee that the energy function itself is represented accurately. When modeled by a conventional multilayer perceptron, the Hamiltonian inherits the spectral bias of coordinate-based neural representations, favoring smooth over fine-scale structure. These fine-scale errors can be amplified by differentiation when computing the dynamics. To address this limitation, we introduce Local Multiscale Hamiltonian Neural Network (LM-HNN), which restructures the Hamiltonian according to its physical interaction structure. LM-HNN decomposes the energy into kinetic and potential components and represents them as sums of shared local functions over multiple interaction neighborhoods. This replaces a single high-dimensional approximation problem with lower-dimensional, reusable interaction laws, making fine-scale energy structure easier to resolve while preserving Hamiltonian dynamics. Across low-dimensional, interacting, and fast--slow systems, LM-HNN consistently achieves more accurate Hamiltonian derivatives and long-horizon rollouts than competing structure-preserving models, while substantially extending the effective spectral range of the learned energy. The same structured representation also generalizes more reliably to energy levels not observed during training. These results show that accuracy depends on the Hamiltonian representation as well as its structure, and that incorporating locality, decomposition, and shared interaction laws provides a simple and effective route to more accurate and generalizable Hamiltonian learning.

%% file: sections/1_introduction.tex
\section{Introduction}
\label{sec:introduction}

Learning Hamiltonian dynamics can be challenging when the underlying energy function contains fine-scale variations. Hamiltonian neural networks (HNNs) obtain a vector field by differentiating a learned scalar energy $H_{\theta}(\bq,\bp)$, making the quality of this representation central to prediction~\citep{greydanus2019hamiltonian}. A standard HNN uses a global Hamiltonian representation, where a single network maps the full phase-space state to the total energy. Standard HNNs therefore inherit the spectral bias~\citep{rahaman2019spectral,tancik2020fourier} of coordinate networks, learning low-frequency energy variations more readily than fine-scale structure. Since the energy is differentiated, unresolved high-frequency components can cause substantial vector-field errors even when their energy amplitude is small, and these errors can accumulate during rollout~\citep{sitzmann2020siren,gruver2022deconstructing}.

Efforts to address spectral bias have improved the recovery of fine-scale signals by changing coordinate encodings and activation functions~\citep{tancik2020fourier,sitzmann2020siren}. These modifications change which frequencies a network learns efficiently, making them natural candidates for improving Hamiltonian approximation. Hamiltonian learning has also benefited from decomposing the energy across temporal scales~\citep{li2026fshnn}. However, improving frequency resolution or separating timescales does not directly encode how the energy is assembled from physical interactions. A global Hamiltonian model must still infer which state variables interact, how their energy contributions combine, and which interaction laws recur across the system.

Encoding these interactions explicitly allows fine-scale energy variation to be learned through local functions, each depending on only a small subset of the state variables. Many mechanical systems admit a kinetic–potential decomposition~\citep{Goldstein2002ClassicalMechanics,marion2013classical}, and finite-range potentials can be represented as sums of local energy contributions, a principle established in neural potential-energy models~\citep{behler2007generalized}. Repeated instances of the same interaction provide supervision for a shared local function. This suggests a mechanism for mitigating spectral bias by reducing the dimensionality of the functions being learned and concentrating shared capacity on the underlying interaction laws. The same factorization can support generalization when globally unseen states contain local configurations similar to those encountered during training.

Building on these insights, we introduce the Local Multiscale Hamiltonian Neural Network (LM-HNN). The model learns shared local functions for kinetic and potential energies, then sums their outputs to form the total Hamiltonian. Evaluating these functions over several neighborhood radii accommodates different interaction ranges without committing to a single radius in advance. This local factorization targets spectral bias by concentrating each network’s capacity on a lower-dimensional interaction law and its fine-scale energy variations. Aggregating the learned contributions also supports generalization to unseen energy levels when new total energies arise from combinations of local configurations represented during training. The assembled scalar Hamiltonian generates the dynamics through Hamilton’s equations, preserving Hamiltonian structure by construction. For systems with repeated interactions and fixed neighborhood sizes, parameter sharing also makes the parameter count independent of system size. We evaluate these properties through spectral diagnostics, tests at unseen energy levels, and prediction accuracy across six conservative systems. Comparisons with structure-preserving baselines and spectral interventions show that LM-HNN achieves lower derivative and rollout errors, while ablations identify locality and kinetic–potential decomposition as the main sources of these gains.

Concretely, our contributions are as follows:
\begin{itemize}[leftmargin=6mm,topsep=0pt,itemsep=0pt]
    \vspace{-1mm}
    \item We analyze the spectral bias of conventional HNNs, and show that it is measurable directly on the learned Hamiltonian rather than only from prediction error.
    \item We propose LM-HNN, which combines kinetic and potential decomposition with shared local energy functions evaluated across multiple neighborhood scales. The representation extends the spectral range over which the Hamiltonian is resolved and improves derivative and rollout accuracy against structure-preserving baselines.
    \item We show that LM-HNN generalizes to energy levels outside the training range, where the baselines lose accuracy, supporting more accurate dynamics beyond the training-energy range.
\end{itemize}

%% file: sections/2_related_work.tex
\section{Related Work}
\label{sec:related}

\textbf{Structure-preserving learning of dynamical systems.}
Data-driven modeling of dynamical systems spans sparse system identification~\citep{brunton2016discovering}, continuous-time neural models~\citep{chen2018neural}, and neural operators for spatially distributed systems~\citep{li2020fourier,lu2021learning}. For conservative dynamics, a related line of work incorporates mechanics directly into the model class. Hamiltonian neural networks learn a scalar energy and derive the vector field through Hamilton’s equations~\citep{greydanus2019hamiltonian,bertalan2019learning}, while Lagrangian networks parameterize the Lagrangian and recover the corresponding equations of motion~\citep{lutter2019deep,cranmer2020lagrangian}. Other approaches preserve geometric structure through symplectic integration or directly parameterized symplectic maps~\citep{chen2019symplectic,zhong2020symplectic,jin2020sympnets,hairer2006geometric}, with related extensions to non-canonical and PDE settings~\citep{eidnes2024pseudo}. These methods primarily constrain the geometry of the learned dynamics. Our focus is complementary: even with a valid Hamiltonian structure, prediction quality still depends on how accurately the underlying scalar Hamiltonian can be represented.

\textbf{Implicit neural representations and spectral bias.}
Coordinate-based MLPs, or implicit neural representations (INRs), learn continuous signals directly from their coordinates. A key limitation is spectral bias: standard MLPs tend to fit low-frequency components much faster than high-frequency ones~\citep{rahaman2019spectral,tancik2020fourier}. Here, frequency is defined over the coordinate domain rather than time. To improve high-frequency representation, prior work modifies either the representation or the optimization process. Fourier features enrich the input spectrum~\citep{tancik2020fourier}, SIREN uses periodic activations~\citep{sitzmann2020siren}, and methods such as WIRE, FINER, and BACON provide more explicit control over spatial-frequency content and bandwidth~\citep{saragadam2023wire,liu2024finer,lindell2022bacon}. Other approaches instead modify training dynamics, for example through neural-tangent-kernel-based gradient reweighting~\citep{shi2025inductive}. A standard HNN can also be viewed as an INR over phase space. So spectral error is particularly important here for a better approximation of the dynamic system. High-frequency approximation errors can therefore be amplified through differentiation and directly distort the learned vector field. Rather than introducing a generic frequency encoding or activation, we create a new method to approximate the Hamiltonian using its kinetic--potential structure, locality, and repeated interactions, making the target function easier for the underlying coordinate networks to represent.

\textbf{Local and multiscale representations.}
Multiscale representations have a long history in numerical analysis, including multigrid and multiresolution methods that organize information across spatial scales~\citep{mandel1988algebraic,mallat1989multiresolution,hughes1995multiscale}. Similar ideas appear in modern neural representations. NeRF-style models represent continuous spatial fields with shared coordinate networks~\citep{mildenhall2020nerf}, while mip-NeRF and multiresolution hash encodings combine information across multiple spatial scales~\citep{barron2021mipnerf,muller2022instantngp}. LM-HNN adopts this representation principle rather than the rendering objective. It expresses the Hamiltonian through shared local energy functions evaluated over multiple neighborhood radii, reducing the complexity seen by each network while directly encoding finite-range interactions. Sharing the same local energy law across sites further allows the representation to be reused across spatial locations and system sizes. In this sense, local multiscale construction serves as a physics-guided representations for making the Hamiltonian easier to approximate, rather than explicit Fourier decomposition of the dynamics.

%% file: sections/3_background.tex
\section{Background}
\label{sec:background}

\textbf{Hamiltonian Dynamics.}
Hamiltonian mechanics describes a conservative system using generalized coordinates $\bq$, conjugate momenta $\bp$, and a scalar Hamiltonian $H(\bq,\bp)$ representing the total energy~\citep{Goldstein2002ClassicalMechanics,marion2013classical}. With $\bz=(\bq,\bp)$, the dynamics are
\begin{equation}
\dot{\bz}=J\nabla_{\bz}H(\bz), \quad
\text{where}\quad J=\begin{bmatrix}0&I\\-I&0\end{bmatrix},
\label{eq:canonical}
\end{equation}
and the exact flow conserves $H$ and preserves the symplectic structure~\citep{hairer2006geometric}. Hamiltonian neural networks (HNNs)~\citep{greydanus2019hamiltonian,toth2019hamiltonian} approximate $H$ with a differentiable model $H_\theta$, yielding vector field $J\nabla H_\theta$. Given $M$ observed state–derivative pairs, the model is trained by minimizing
\begin{equation}
    \cL_{\mathrm{deriv}} = \frac{1}{M}\sum\nolimits_{i=1}^{M}\left\lVert J\nabla H_\theta(\bz_i)-\dot{\bz}_i\right\rVert_2^2.
    \label{eq:deriv-loss}
\end{equation}

\textbf{Implicit Neural Representation.}  A standard HNN can be viewed as an implicit neural representation (INR) of the Hamiltonian over phase space, where a coordinate network maps the continuous state $\bz$ to the scalar energy $H_\theta(\bz)$, whose gradient determines the dynamics. Hamiltonian mechanics constrains the structure of the vector field, while the neural representation determines how accurately the energy and its derivatives are approximated. Preserving Hamiltonian structure alone therefore does not guarantee accurate dynamics. The energy residual $\delta H=H_\theta-H$ affects the learned vector field through its gradient, giving
\begin{equation}
    \left\lVert J\nabla H_\theta(\bz)-\dot{\bz}\right\rVert_2 = \left\lVert\nabla\delta H(\bz)\right\rVert_2.\label{eq:field-error}
\end{equation}
where the equality follows from $J^\top J=I$. A constant offset has no effect, whereas a small but rapidly varying residual can produce a large vector-field error. Conventional HNNs represent the total energy using a single network over the full state, leaving kinetic–potential decomposition, locality, and repeated interactions to be learned implicitly from data.


\textbf{Spectral Bias of Coordinate-Based Networks.}
Coordinate-based MLPs exhibit spectral bias, typically learning low-frequency target components more rapidly than high-frequency ones under gradient-based optimization~\citep{rahaman2019spectral,xu2020frequency}. Here, frequency refers to variation over phase-space coordinates rather than temporal frequency along a trajectory. The neural tangent kernel (NTK) provides an explanation through the learning rates of different residual modes~\citep{jacot2018ntk}. Under linearized gradient flow for squared-error function fitting, the component of the residual $r=f_\theta-f$ along eigenfunction $v_{j}$ of the fixed NTK operator $\Theta$ evolves as
\begin{equation}
    \langle v_j,r(t)\rangle=e^{-\lambda_j t}\langle v_j,r(0)\rangle,\label{eq:ntk-modes}
\end{equation}
where $\lambda_{j}$ is the corresponding eigenvalue and $t$ denotes optimization time. Higher-frequency components are often associated with smaller eigenvalues in conventional fully connected networks, explaining their slower convergence~\citep{basri2019convergence,cao2021towards,xu2020frequency}. Fourier features~\citep{tancik2020fourier} and periodic activations~\citep{sitzmann2020siren} modify this spectral behavior and improve the recovery of fine-scale variation.

For HNNs, unresolved high-frequency structure is particularly consequential because the learned energy is differentiated to obtain the dynamics. For the energy residual $\delta H$, Parseval’s identity gives
\begin{equation}
  \int \|\nabla \delta H(\mathbf{z})\|_2^2\,\mathrm d\mathbf{z} = \int \|\mathbf{k}\|_2^2 \bigl|\widehat{\delta H}(\mathbf{k})\bigr|^2\,\mathrm d\mathbf{k}.\label{eq:parseval-grad}
\end{equation}
Higher-wavenumber components therefore receive greater weight in the vector-field error, even when their energy amplitude is small. Derivative supervision emphasizes these errors during training, but does not by itself guarantee their accurate recovery~\citep{wang2022when}. This motivates improving how the Hamiltonian is represented. LM-HNN uses shared local energy functions with lower-dimensional inputs, allowing repeated interactions to be learned through a common representation. The next section develops this construction while retaining Hamiltonian structure.

%% file: sections/4_method.tex
\section{Method}
\label{sec:method}

We design the Local Multiscale Hamiltonian Neural Network (LM-HNN) to improve the approximation of fine-scale energy variations by representing the Hamiltonian through lower-dimensional energy functions. We first decompose the Hamiltonian into kinetic and potential components (Sec.~\ref{sec:method-decomposition}), then represent each through shared local functions over multiple neighborhood scales (Sec.~\ref{sec:method-local}). We discuss how this lower-dimensional representation supports the recovery of fine-scale energy structure and generalization to unseen energy levels (Sec.~\ref{sec:method-representation}). The dynamics follow from the resulting scalar Hamiltonian through Hamilton's equations, preserving Hamiltonian structure.

\subsection{Energy Decomposition and Local Neighborhoods}
\label{sec:method-decomposition}

Following the decomposition of the Hamiltonian into kinetic and potential energy~\citep{Goldstein2002ClassicalMechanics}, we parameterize the two components using separate functions,
\begin{equation}
    H_\theta(\bq,\bp)=T_{\varphi}(\bq,\bp)+V_{\phi}(\bq),\label{eq:tv-split}
\end{equation}
where $T_{\varphi}$ and $V_{\phi}$ denote the kinetic and potential energies respectively. The potential energy depends only on the configuration, while the kinetic energy may depend on both coordinates and momenta, accommodating configuration-dependent inertia and momentum coupling. This separation allows each component to use a parameterization suited to its physical structure. For kinetic energies that are quadratic and diagonal in momentum, we encode the quadratic dependence explicitly and learn the inverse-mass coefficients.

To describe the interaction structure within these components, let $\mathcal N_r(i)$ denote the interaction neighborhood of degree of freedom $i$ at radius $r$. The corresponding local variables are
\begin{equation}
    \bq_i^r=\bq_{\mathcal N_r(i)}, \qquad \bp_i^r=\bp_{\mathcal N_r(i)}.\label{eq:local-neighborhood}
\end{equation}
These neighborhoods specify which variables enter each local energy function. They are determined by the system’s interaction structure, with increasing $r$ providing broader local context. On the chains, $\mathcal N_r(i)$ is a window of bond strains for the potential and a window of sites for the kinetic energy; the inputs on every system are listed in Appendix~\ref{app:training-protocol}. The next subsection uses these variables to construct shared energy functions and combine their contributions across neighborhood scales.

\subsection{Local Multiscale Energy Parameterization}
\label{sec:method-local}

\textbf{Potential Energy.} We represent the potential energy as a weighted sum of shared local contributions,
\begin{equation}
    V_\theta(\bq) = \sum\nolimits_{r\in\mathcal R}\pi_r^V \sum\nolimits_{i\in\mathcal I_V}v_{\theta,r}\left(\bq_i^r\right),\label{eq:local-potential}
\end{equation}
where $\mathcal R$ is the set of neighborhood radii and $\mathcal I_V$ is the index set of local contributions to the potential energy. The scale weights are then parameterized as
\begin{equation}
    \pi_r^V = \frac{\exp(\eta_r^V)}{\sum_{r'\in\mathcal R}\exp(\eta_{r'}^V)},\label{eq:potential-scale-weight}
\end{equation}
where $\eta_{r}^{V}$ are learnable scale logits.
At each radius, the shared network $v_{\theta,r}$ learns a common interaction law across neighborhoods with the same local structure. Combining contributions across scales incorporates both short- and longer-range interaction context.

\textbf{Kinetic Energy.} For kinetic energies that are quadratic and diagonal in momentum, we use
\begin{equation}
    T_\theta(\bq,\bp) = \frac{1}{2}\sum\nolimits_{i=1}^{N}m^{-1}_{\theta,i}(\bq)p_i^2,\label{eq:local-kinetic}
\end{equation}
with locally parameterized inverse-mass coefficients
\begin{equation}
    m^{-1}_{\theta,i}(\bq) = \epsilon+\operatorname{softplus}\left(\sum\nolimits_{r\in\mathcal R}\pi_r^Ta_{\theta,r}\left(\bq_i^r\right)\right).\label{eq:local-mass}
\end{equation}
At each radius $r$, the same function $a_{\theta,r}$ is applied to the local coordinates around every site, while $\pi_{r}^{T}$ weights their contributions across radii. This parameterization explicitly encodes the quadratic dependence on momentum and learns the configuration-dependent inverse masses.

When momentum coupling is present, we instead use
\begin{equation}
    T_\theta(\bq,\bp) = \sum\nolimits_{r\in\mathcal R}\pi_r^T \sum\nolimits_{i\in\mathcal I_T}t_{\theta,r}\left(\bq_i^r,\bp_i^r\right),\label{eq:kinetic-sum}
\end{equation}
where $\mathcal{I}_{T}$ is the index set of local terms used to construct the learned kinetic energy. The shared functions $t_{\theta,r}$ depend jointly on the coordinates and momenta within each neighborhood, allowing them to represent local momentum coupling. These constructions assemble the Hamiltonian from shared, lower-dimensional energy functions matched to the system’s interaction structure.
\subsection{Spectral Resolution and Energy Generalization}
\label{sec:method-representation}

For systems with repeated local interactions, local factorization reduces the dimensionality of Hamiltonian approximation by expressing the total energy as a sum of shared local functions,
\begin{equation}
    H_\theta(\bz) = \sum\nolimits_i h_\theta\left(\bz_{\mathcal N(i)}\right), \label{eq:shared-local-hamiltonian}
\end{equation}
where each local function receives only the variables involved in an interaction. Fine-scale variation is therefore represented directly in the coordinates on which it depends. Parameter sharing aggregates the contributions of repeated interactions, such that
\begin{equation}
    \nabla_\theta H_\theta(\bz) = \sum\nolimits_i \nabla_\theta h_\theta\left(\bz_{\mathcal N(i)}\right). \label{eq:shared-gradient}
\end{equation}
These interactions are therefore captured by a common set of parameters, allowing the same local energy law to be learned across the system. Lower-dimensional inputs and repeated supervision can thereby support the recovery of fine-scale Hamiltonian structure.

The same factorization provides a mechanism for generalization to unseen energy levels. Let $\mathcal S_{\mathrm{train}}$ denote the training region in phase space and let $P_i\bz=\bz_{\mathcal N(i)}$ be the local projection. An unseen state may lie outside $\mathcal S_{\mathrm{train}}$ while its local configurations remain within or near the projected domains $P_i\mathcal S_{\mathrm{train}}$. Under this condition, the representation
\begin{equation}
    H_\theta(\bz) = \sum\nolimits_i h_\theta\left(P_i\bz\right)\label{eq:local-energy-generalization}
\end{equation}
evaluates new combinations of familiar local configurations, reducing the need for extrapolation within each local function. The kinetic–potential decomposition further reduces the approximation burden. When the diagonal quadratic kinetic form is used, momentum dependence is encoded explicitly, leaving the networks to learn configuration-dependent inverse-mass coefficients and interaction potentials. These functions can then be reused across different states and energy levels.

%% file: sections/5_experiments.tex
\section{Experiments}
\label{sec:experiments}

We evaluate the proposed representation through the learned Hamiltonian, its vector field, and the resulting rollouts. We first examine spectral bias and compare with existing mitigation techniques, then assess prediction accuracy within and beyond the training energy range across six conservative systems against structure-preserving baselines. Ablations examine which components contribute to performance and how their benefits vary with interaction structure, system size, and nonlinearity.

\subsection{Dynamical Systems and Baselines}
\label{sec:exp-systems}


\textbf{Dynamical systems.} We evaluate LM-HNN on pendulum and double pendulum,  Fermi–Pasta–Ulam-Tsingou (FPUT) chain~\citep{fpu1955,benettin2013fermi} under nearest-neighbor, Toda~\citep{toda1967vibration}, and Morse~\citep{morse1929diatomic} potential, and the fast-slow Hamiltonian oscillator described in \citet{duruisseaux2023approximation}. These six conservative systems span low-dimensional and many-body dynamics, separable and non-separable kinetic energies, different interaction laws, and separated timescales, providing a controlled setting for studying representation accuracy, spectral bias mitigation, and energy generalization. We include detailed definitions of these systems in {Appendix}~\ref{app:systems}.

To study spectral bias in a controlled setting, we introduce a FPUT variant with perturbed potential
\begin{equation}
    \phi_k(s)=\phi_{\mathrm{FPUT}}(s)+\frac{a}{k}(1-\cos ks),\label{eq:perturbed-fput}
\end{equation}
by adding force term $a\sin(ks)$. We fix $a=0.5$ and vary $k$, thereby increasing the frequency of the Hamiltonian variation while keeping the force amplitude fixed.





\textbf{Baselines.} HNN~\citep{greydanus2019hamiltonian} models the Hamiltonian using a single network, providing the direct comparison with our local decomposition under the same Hamiltonian formulation. SympNet~\citep{jin2020sympnets} learns a symplectic discrete-time map, testing whether comparable accuracy can be achieved by preserving the geometry of evolution without explicitly learning the energy. FS-HNN~\citep{li2026fshnn} learns separate Hamiltonian branches for temporal frequency bands, allowing us to compare decomposition by timescale and decomposition by physical interaction. The HNN and the variants of Table~\ref{tab:interv} receive the same inputs as LM-HNN (Appendix~\ref{app:training-protocol}).



\subsection{Spectral Bias of the Learned Hamiltonian}
\label{sec:exp-bias}

\begin{figure}[b]
    \vspace{-0.1in}
    \centering
    \includegraphics[width=\linewidth]{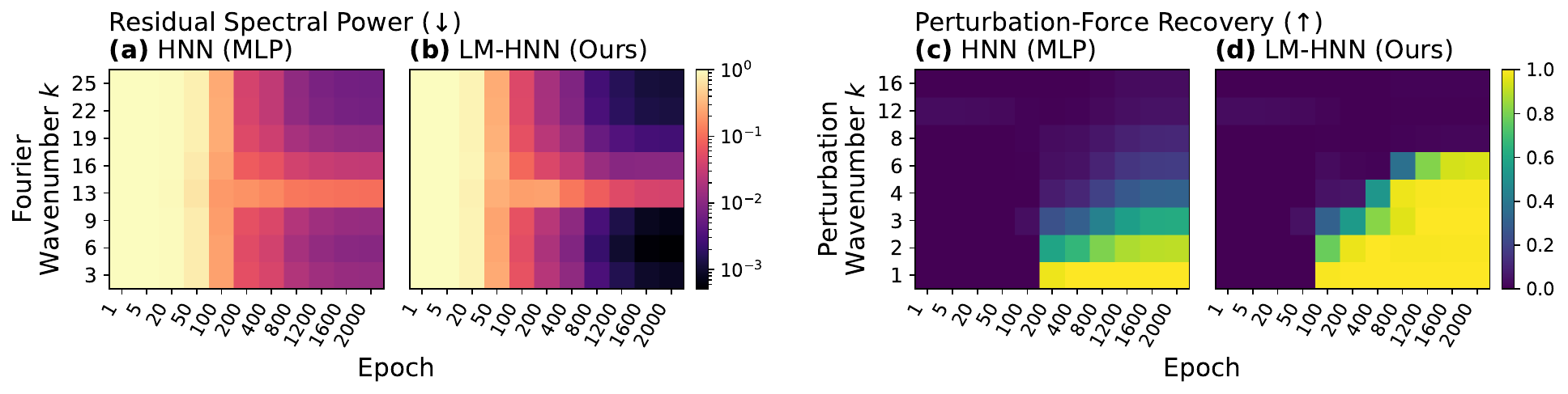}
    \vspace{-0.32in}
    \caption{\textbf{Spectral bias in the learned Hamiltonian on the FPUT chain ($\mathbf{N=8}$).} \textbf{(a, b)} Residual Hamiltonian spectral power during training on the unperturbed FPUT chain, normalized by the target power at each Fourier wavenumber $k$. \textbf{(c, d)} Recovery of the added force component during training on the perturbed chain, where $k$ now denotes the prescribed perturbation wavenumber, varied while keeping the force amplitude fixed. At matched parameter counts, LM-HNN reduces higher-wavenumber residuals more rapidly and recovers finer imposed interactions.}
    \label{fig:spectral_bias_combined}
\end{figure}
\begin{table}[t]
    \centering
    \caption{\textbf{Baselines for mitigating spectral bias.} Derivative and rollout MSE for FPUT and pendulum for baseline mitigation techniques applied to HNN at matched parameter count. The structural approach of LM-HNN outperforms both Fourier features and sinusoidal activations.}
    \label{tab:interv}
    \setlength{\tabcolsep}{4pt}
    \vspace{0.03in}
    \input{tables/table_interv}
    \vspace{-0.15in}
\end{table}

Fine-scale Hamiltonian components can contribute substantially to the vector field even when their energy amplitudes are small, making their recovery important for accurate dynamics. We therefore test whether the proposed local decomposition makes these components easier to learn by examining Hamiltonian spectral errors during training on the unperturbed FPUT chain. Figure~\ref{fig:spectral_bias_combined} (a) and (b) tracks the spectral power of the difference between the learned and true Hamiltonians during training. HNN and LM-HNN both learn low-wavenumber components first, but LM-HNN reduces higher-wavenumber residuals more rapidly at matched parameter counts and within the same training budget. Lower Hamiltonian residuals and vector-field errors show that the improvement is present in the learned energy and its derivatives before numerical integration (Table~\ref{tab:bias-networks} and Figure~\ref{fig:spectra}, Appendix~\ref{app:spectral-diagnostics}).

To test whether the improved recovery on the unperturbed chain extends to finer energy variations, we use the perturbed FPUT potential in Equation~\ref{eq:perturbed-fput}. We increase the perturbation wavenumber $k$, which controls how rapidly the added force $a\sin(ks)$ varies with bond displacement $s$, while keeping its amplitude $a$ fixed. Figure~\ref{fig:spectral_bias_combined} (c) and (d) shows that HNN’s recovery of this added force component deteriorates from approximately $k=3$, whereas LM-HNN reaches nearly complete recovery through $k=6$. The improved force recovery is accompanied by lower rollout MSE (Table~\ref{tab:testbed}), reflecting more accurate state predictions over time, and lower peak error (Table~\ref{tab:testbed-peak}), indicating better agreement with the reference trajectory amplitudes (Appendix~\ref{app:perturbed-fpu}).

To test whether these gains can be achieved without decomposing the Hamiltonian, we compare LM-HNN with HNNs using Fourier features~\citep{tancik2020fourier} or sinusoidal activations~\citep{sitzmann2020siren}. Table~\ref{tab:interv} reports derivative and rollout errors at matched parameter counts. In the tested configurations, Fourier features increase both errors. Sinusoidal activations can improve derivative accuracy, but these gains do not extend to rollouts from unseen initial conditions. LM-HNN improves both metrics over HNN on FPUT and pendulum and achieves the lowest rollout errors, indicating that local energy decomposition provides benefits not reproduced by these spectral interventions.

\subsection{Generalization to Unseen Energy Levels}
\label{sec:exp-generalization}

\begin{table}[b]
    \vspace{-0.12in}
    \caption{\textbf{Normalized rollout MSE at unseen energy levels.} The FPUT chains are trained on $E/N\in[0.05,2.0]$ and tested at $2.5$ and $3.0$ per site; the pendulum is trained on $E\in[0.05,4]$ and tested at $6$ and $8$; and the double pendulum is trained on $E\in[0.5,5]$ and tested at $7$ and $9$. The fast--slow row is a scaled-initial-condition test: the test states are multiplied by $1.5$, which places 9 of the 30 above the largest training energy of $2.48$ (Appendix~\ref{app:systems}). Baseline methods exhibit increased rollout error at unseen energy levels whilst LM-HNN maintains low error.}
    \label{tab:generalization}
    \vspace{0.03in}
    \centering
    \setlength{\tabcolsep}{4pt}

\input{tables/table_generalization}
\end{table}

\begin{figure}[t]
    \centering
    \includegraphics[width=\linewidth]{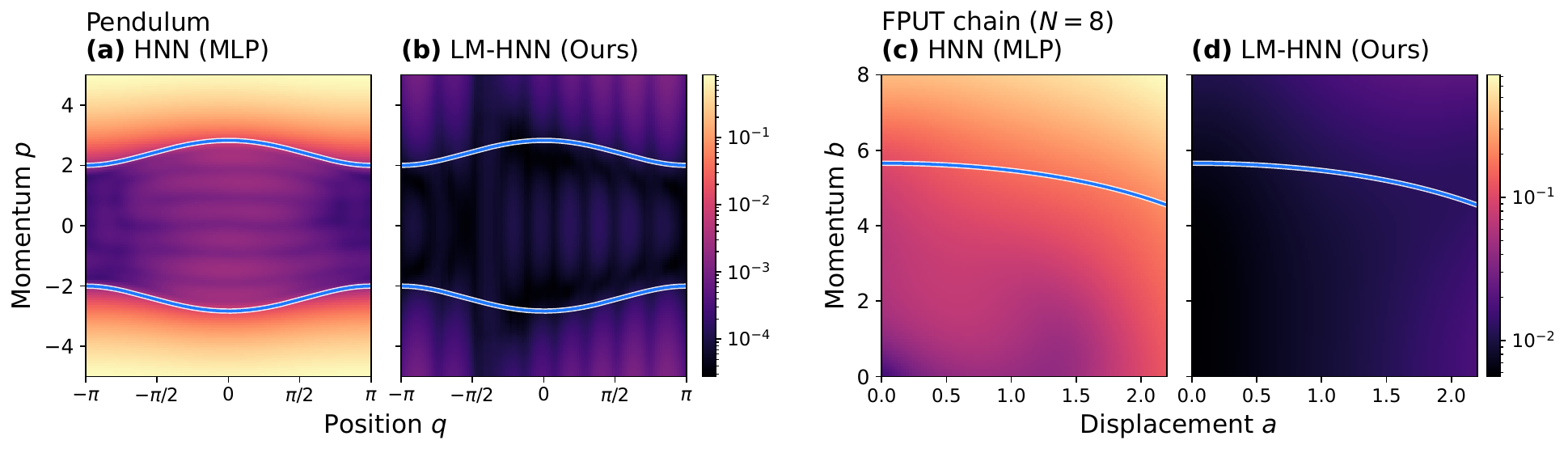}
    \vspace{-0.3in}
    \caption{\textbf{Vector-field accuracy beyond the training-energy range.} Vector-field error $\lVert J\nabla H_\theta(\bz)-\dot{\bz}\rVert_2$, shown as a local root-mean-square for \textbf{(a, b)} the pendulum phase plane and \textbf{(c, d)} the FPUT chain ($N=8$) on the slice $\bq=a\mathbf u$, $\bp=b\mathbf v$ along fixed random unit directions. Blue curves mark the maximum training energy, and each pair shares a logarithmic color scale. LM-HNN maintains lower errors beyond this boundary, supporting generalization to unseen energy levels.}
    \label{fig:energy-surface}
    \vspace{-0.15in}
\end{figure}

Having established improved recovery of fine-scale Hamiltonian structure, we next examine whether LM-HNN’s learned energy components support accurate dynamics beyond the training energy range. We evaluate the trained models from higher-energy initial conditions for the pendulum and chain systems and use a scaled-initial-condition test for the fast--slow oscillator, multiplying its test states by $1.5$, which places 9 of the 30 above the largest training energy (Appendix~\ref{app:systems}); the settings are summarized in Table~\ref{tab:generalization}.

Figure~\ref{fig:energy-surface} compares errors in the predicted state derivatives for the pendulum and FPUT chain, with blue curves marking the maximum training energy. HNN’s errors grow substantially beyond this boundary, whereas LM-HNN maintains lower errors and exhibits a smaller increase. 

Table~\ref{tab:generalization} shows that this advantage extends to rollout prediction at unseen energies. The baselines generally reach normalized errors of order one, comparable to the variance of the reference trajectories, whereas LM-HNN’s mean error remains below $0.2$ in every row, including the scaled-initial-condition test on the fast--slow oscillator. The separation is especially large for the pendulum systems, where LM-HNN retains very small prediction errors. LM-HNN’s rollout errors increase on the chains, where higher energies produce larger bond strains, but remain substantially below the baselines across FPUT, FPUT-Toda, and FPUT-Morse.

\begin{table}[b]
    \vspace{-0.12in}
    \caption{\textbf{Rollout and derivative MSE across six conservative systems.} SympNet learns a discrete symplectic map and therefore does not provide a continuous vector field for the derivative comparison. LM-HNN achieves the lowest rollout and derivative error on every system, showing consistent improvement across low-dimensional, lattice, and fast--slow dynamics.}
    \label{tab:systems}
    \vspace{0.03in}
    \centering
    \setlength{\tabcolsep}{4pt}

\input{tables/table_systems}
\end{table}
\begin{table}[t]
    \centering
    \caption{\textbf{Structural ablations of LM-HNN.}
    Normalized rollout MSE on the FPUT and Toda chains ($N=8$).
    Locality and kinetic--potential decomposition improve accuracy,
    while weight sharing reduces model size without a consistent
    accuracy penalty. Full metrics are reported in
    Appendix~\ref{app:ablation}.}
    \label{tab:ablation-main}
    \vspace{0.03in}
    \begin{tabular}{@{}lr*{2}{r@{}l}@{}}
        \toprule
        Ablation Variant & Param. & \multicolumn{2}{c}{FPUT} & \multicolumn{2}{c}{FPUT-Toda} \\
        \midrule
        LM-HNN
        & 26,892
        & $(4.9{\scriptstyle\pm0.6})$&${\scriptstyle\times10^{-3}}$
        & $\mathbf{(6.7{\scriptstyle\pm4.7})}$&$\mathbf{{\scriptstyle\times10^{-3}}}$ \\
        w/o kinetic--potential split
        & 26,736
        & $(3.8{\scriptstyle\pm0.6})$&${\scriptstyle\times10^{-2}}$
        & $(7.1{\scriptstyle\pm1.2})$&${\scriptstyle\times10^{-2}}$ \\
        w/o locality
        & 26,932
        & $(5.5{\scriptstyle\pm0.5})$&${\scriptstyle\times10^{-2}}$
        & $(3.8{\scriptstyle\pm1.0})$&${\scriptstyle\times10^{-1}}$ \\
        w/o weight sharing
        & 228,537
        & $\mathbf{(3.5{\scriptstyle\pm0.8})}$&$\mathbf{{\scriptstyle\times10^{-3}}}$
        & $(1.9{\scriptstyle\pm0.2})$&${\scriptstyle\times10^{-2}}$ \\
        \bottomrule
    \end{tabular}
    \vspace{-0.12in}
\end{table}

On the pendulum the angular encoding is shared with the HNN, so it does not explain the gap; energies above the range are faster rotations with larger $|p|$, which LM-HNN evaluates through its fixed quadratic form $\tfrac12 m_\theta^{-1}(q)\,p^2$ while the HNN must extrapolate an unconstrained dependence on $p$. The chain results support the usefulness of shared local potentials over a wider range of configurations. Table~\ref{tab:diag} further shows that smaller errors between the learned and true Hamiltonians on validation states accompany lower rollout errors at unseen energies (Appendix~\ref{app:error-generalization}), connecting Hamiltonian approximation quality to generalization.

\subsection{Accuracy of the Learned Hamiltonian Across Systems}
\label{sec:exp-accuracy}


The spectral and unseen-energy results motivate testing whether local energy decomposition improves prediction across different kinetic couplings, interaction potentials, and timescales. We evaluate all six systems from unseen initial conditions within the training distribution, using comparable parameter budgets (Table~\ref{tab:systems-params}, Appendix~\ref{app:accuracy}).

Table~\ref{tab:systems} shows that LM-HNN achieves the lowest mean rollout error across all six systems, demonstrating consistent benefits across different kinetic couplings, interaction laws, and timescales. The accompanying reduction in validation derivative errors (Table~\ref{tab:systems}) shows that these gains extend to the underlying vector fields, connecting more accurate learned dynamics to better trajectory predictions.

Figure~\ref{fig:rollout-case} illustrates the resulting trajectory behavior through reduced phase and amplitude errors on the chains and more accurate tracking of the slow drift and fast oscillation phase in the fast–slow system (Appendix~\ref{app:accuracy}).




\subsection{Ablations}
\label{sec:exp-ablation}

To identify which structural choices drive LM-HNN's accuracy gains, we ablate its components on the FPUT and FPUT-Toda chains. Removing locality produces the largest increase in rollout error, followed by removing the kinetic--potential split (Table~\ref{tab:ablation-main}). Both variants retain Hamiltonian structure and similar parameter counts, supporting the importance of organizing the energy approximation through local functions and separate energy components. Removing weight sharing increases the parameter count by approximately $8.5\times$ without a consistent accuracy gain. Locality and energy decomposition therefore contribute most directly to accuracy, while sharing represents repeated interactions efficiently.

The receptive-field study clarifies the role of multiple scales (Table~\ref{tab:radius}, Appendix~\ref{app:ablation}). A radius-zero neighborhood contains a single bond strain and performs best on the nearest-neighbor chain, but cannot represent the added second-neighbor coupling, for which the best single radius is $\mathcal R=\{1\}$. The multiscale set $\mathcal R=\{0,1,2\}$ accommodates both interaction ranges and achieves the lowest derivative error on the two-scale chain, while matched single radii yield lower rollout errors. The kinetic-head comparison similarly favors diagonal quadratic forms for uncoupled momenta, local sums for finite-range momentum coupling, and a dense representation for global coupling (Table~\ref{tab:locality}). 

The chain-length sweep further supports the shared local representation (Table~\ref{tab:nscale}). From $N=4$ to $N=16$, LM-HNN retains the same parameter count and keeps mean rollout error below $10^{-2}$, while its advantage over parameter-matched HNNs widens. This is consistent with learning local interaction laws whose complexity does not grow with the number of repeated interactions. The accuracy advantage also persists across the tested interaction nonlinearities (Table~\ref{tab:elastic}).

%% file: tables/table_interv.tex
\begin{tabular}{l*{4}{r@{}l}}
\toprule
& \multicolumn{4}{c}{FPUT} & \multicolumn{4}{c}{Pendulum} \\
\cmidrule(lr){2-5}\cmidrule(lr){6-9}
Model & \multicolumn{2}{c}{Deriv.\ MSE} & \multicolumn{2}{c}{Rollout MSE} & \multicolumn{2}{c}{Deriv.\ MSE} & \multicolumn{2}{c}{Rollout MSE} \\
\midrule
HNN & $(3.6{\scriptstyle\pm0.3})$&${\scriptstyle\times10^{-2}}$ & $(2.0{\scriptstyle\pm0.1})$&${\scriptstyle\times10^{-1}}$ & $(1.9{\scriptstyle\pm0.7})$&${\scriptstyle\times10^{-6}}$ & $(2.7{\scriptstyle\pm0.6})$&${\scriptstyle\times10^{-5}}$ \\
Fourier ($\sigma$=0.5) & $(1.8{\scriptstyle\pm0.0})$&${\scriptstyle\times10^{-1}}$ & $(4.1{\scriptstyle\pm0.4})$&${\scriptstyle\times10^{-1}}$ & $(1.2{\scriptstyle\pm0.2})$&${\scriptstyle\times10^{-5}}$ & $(3.0{\scriptstyle\pm3.2})$&${\scriptstyle\times10^{-2}}$ \\
Fourier ($\sigma$=2) & $(1.8{\scriptstyle\pm0.1})$&${\scriptstyle\times10^{0}}$ & $(3.7{\scriptstyle\pm0.6})$&${\scriptstyle\times10^{0}}$ & $(3.7{\scriptstyle\pm0.2})$&${\scriptstyle\times10^{-4}}$ & $(9.3{\scriptstyle\pm3.7})$&${\scriptstyle\times10^{-2}}$ \\
SIREN ($\omega_0$=10) & $(8.1{\scriptstyle\pm0.4})$&${\scriptstyle\times10^{-1}}$ & $(3.3{\scriptstyle\pm0.3})$&${\scriptstyle\times10^{0}}$ & $(1.9{\scriptstyle\pm0.2})$&${\scriptstyle\times10^{-7}}$ & $(3.3{\scriptstyle\pm2.8})$&${\scriptstyle\times10^{-4}}$ \\
SIREN ($\omega_0$=30) & $(3.0{\scriptstyle\pm0.1})$&${\scriptstyle\times10^{0}}$ & $(5.4{\scriptstyle\pm0.5})$&${\scriptstyle\times10^{0}}$ & $(1.6{\scriptstyle\pm1.1})$&${\scriptstyle\times10^{-5}}$ & $(1.4{\scriptstyle\pm0.0})$&${\scriptstyle\times10^{0}}$ \\
\midrule
LM-HNN & $\bm{(7.7{\scriptstyle\pm1.4})}$&$\bm{{\scriptstyle\times10^{-4}}}$ & $\bm{(4.9{\scriptstyle\pm0.6})}$&$\bm{{\scriptstyle\times10^{-3}}}$ & $\bm{(7.4{\scriptstyle\pm5.6})}$&$\bm{{\scriptstyle\times10^{-9}}}$ & $\bm{(4.1{\scriptstyle\pm2.5})}$&$\bm{{\scriptstyle\times10^{-8}}}$ \\
\bottomrule
\end{tabular}

%% file: tables/table_generalization.tex
\begin{tabular}{l*{4}{r@{}l}}
\toprule
System & \multicolumn{2}{c}{HNN} & \multicolumn{2}{c}{SympNet} & \multicolumn{2}{c}{FS-HNN} & \multicolumn{2}{c}{LM-HNN} \\
\midrule
Pendulum & $(1.3{\scriptstyle\pm0.2})$&${\scriptstyle\times10^{0}}$ & $(1.0{\scriptstyle\pm0.6})$&${\scriptstyle\times10^{0}}$ & $(6.1{\scriptstyle\pm1.2})$&${\scriptstyle\times10^{-1}}$ & $\bm{(2.7{\scriptstyle\pm2.4})}$&$\bm{{\scriptstyle\times10^{-8}}}$ \\
Double Pendulum & $(1.7{\scriptstyle\pm0.6})$&${\scriptstyle\times10^{0}}$ & $(3.5{\scriptstyle\pm0.6})$&${\scriptstyle\times10^{0}}$ & $(2.2{\scriptstyle\pm0.2})$&${\scriptstyle\times10^{0}}$ & $\bm{(8.1{\scriptstyle\pm5.3})}$&$\bm{{\scriptstyle\times10^{-6}}}$ \\
FPUT & $(1.8{\scriptstyle\pm0.1})$&${\scriptstyle\times10^{0}}$ & $(1.8{\scriptstyle\pm0.1})$&${\scriptstyle\times10^{0}}$ & $(2.2{\scriptstyle\pm0.1})$&${\scriptstyle\times10^{0}}$ & $\bm{(6.5{\scriptstyle\pm1.1})}$&$\bm{{\scriptstyle\times10^{-2}}}$ \\
FPUT-Toda & $(1.7{\scriptstyle\pm0.2})$&${\scriptstyle\times10^{0}}$ & $(1.9{\scriptstyle\pm0.2})$&${\scriptstyle\times10^{0}}$ & $(2.3{\scriptstyle\pm0.4})$&${\scriptstyle\times10^{0}}$ & $\bm{(1.3{\scriptstyle\pm0.4})}$&$\bm{{\scriptstyle\times10^{-2}}}$ \\
FPUT-Morse & $(1.2{\scriptstyle\pm0.1})$&${\scriptstyle\times10^{0}}$ & $(1.5{\scriptstyle\pm0.2})$&${\scriptstyle\times10^{0}}$ & $(1.9{\scriptstyle\pm0.1})$&${\scriptstyle\times10^{0}}$ & $\bm{(4.8{\scriptstyle\pm1.5})}$&$\bm{{\scriptstyle\times10^{-3}}}$ \\
Fast-Slow & $(4.8{\scriptstyle\pm0.8})$&${\scriptstyle\times10^{-1}}$ & $(1.9{\scriptstyle\pm0.4})$&${\scriptstyle\times10^{0}}$ & $(2.7{\scriptstyle\pm0.1})$&${\scriptstyle\times10^{0}}$ & $\bm{(1.4{\scriptstyle\pm0.4})}$&$\bm{{\scriptstyle\times10^{-1}}}$ \\
\bottomrule
\end{tabular}

%% file: tables/table_systems.tex
\begin{tabular}{l*{4}{r@{}l}}
\toprule
System & \multicolumn{2}{c}{HNN} & \multicolumn{2}{c}{SympNet} & \multicolumn{2}{c}{FS-HNN} & \multicolumn{2}{c}{LM-HNN (ours)} \\
\midrule
\multicolumn{9}{l}{\textit{Rollout MSE}} \\
Pendulum & $(2.7{\scriptstyle\pm0.6})$&${\scriptstyle\times10^{-5}}$ & $(2.9{\scriptstyle\pm1.2})$&${\scriptstyle\times10^{-2}}$ & $(6.2{\scriptstyle\pm2.8})$&${\scriptstyle\times10^{-6}}$ & $\bm{(4.1{\scriptstyle\pm2.5})}$&$\bm{{\scriptstyle\times10^{-8}}}$ \\
Double Pendulum & $(4.3{\scriptstyle\pm5.0})$&${\scriptstyle\times10^{-2}}$ & $(1.7{\scriptstyle\pm0.3})$&${\scriptstyle\times10^{0}}$ & $(1.9{\scriptstyle\pm0.9})$&${\scriptstyle\times10^{-2}}$ & $\bm{(3.2{\scriptstyle\pm3.2})}$&$\bm{{\scriptstyle\times10^{-6}}}$ \\
FPUT & $(2.0{\scriptstyle\pm0.1})$&${\scriptstyle\times10^{-1}}$ & $(6.2{\scriptstyle\pm0.6})$&${\scriptstyle\times10^{-2}}$ & $(1.2{\scriptstyle\pm0.4})$&${\scriptstyle\times10^{-2}}$ & $\bm{(4.9{\scriptstyle\pm0.6})}$&$\bm{{\scriptstyle\times10^{-3}}}$ \\
FPUT-Toda & $(7.0{\scriptstyle\pm0.9})$&${\scriptstyle\times10^{-1}}$ & $(1.7{\scriptstyle\pm0.7})$&${\scriptstyle\times10^{-1}}$ & $(2.6{\scriptstyle\pm0.8})$&${\scriptstyle\times10^{-1}}$ & $\bm{(6.7{\scriptstyle\pm4.7})}$&$\bm{{\scriptstyle\times10^{-3}}}$ \\
FPUT-Morse & $(5.1{\scriptstyle\pm0.5})$&${\scriptstyle\times10^{-1}}$ & $(1.3{\scriptstyle\pm0.4})$&${\scriptstyle\times10^{-1}}$ & $(2.1{\scriptstyle\pm0.3})$&${\scriptstyle\times10^{-1}}$ & $\bm{(1.5{\scriptstyle\pm0.4})}$&$\bm{{\scriptstyle\times10^{-3}}}$ \\
Fast-Slow & $(8.7{\scriptstyle\pm3.7})$&${\scriptstyle\times10^{-3}}$ & $(1.3{\scriptstyle\pm0.6})$&${\scriptstyle\times10^{0}}$ & $(2.1{\scriptstyle\pm0.9})$&${\scriptstyle\times10^{-3}}$ & $\bm{(1.8{\scriptstyle\pm0.5})}$&$\bm{{\scriptstyle\times10^{-3}}}$ \\
\midrule
\multicolumn{9}{l}{\textit{Derivative MSE}} \\
Pendulum & $(1.9{\scriptstyle\pm0.7})$&${\scriptstyle\times10^{-6}}$ & \multicolumn{2}{c}{--} & $(5.0{\scriptstyle\pm1.7})$&${\scriptstyle\times10^{-6}}$ & $\bm{(7.4{\scriptstyle\pm5.6})}$&$\bm{{\scriptstyle\times10^{-9}}}$ \\
Double Pendulum & $(1.4{\scriptstyle\pm0.2})$&${\scriptstyle\times10^{-3}}$ & \multicolumn{2}{c}{--} & $(2.2{\scriptstyle\pm0.2})$&${\scriptstyle\times10^{-4}}$ & $\bm{(3.3{\scriptstyle\pm2.2})}$&$\bm{{\scriptstyle\times10^{-6}}}$ \\
FPUT & $(6.0{\scriptstyle\pm0.1})$&${\scriptstyle\times10^{-3}}$ & \multicolumn{2}{c}{--} & $(3.4{\scriptstyle\pm3.3})$&${\scriptstyle\times10^{-3}}$ & $\bm{(5.6{\scriptstyle\pm0.9})}$&$\bm{{\scriptstyle\times10^{-4}}}$ \\
FPUT-Toda & $(1.5{\scriptstyle\pm0.1})$&${\scriptstyle\times10^{-2}}$ & \multicolumn{2}{c}{--} & $(2.7{\scriptstyle\pm0.1})$&${\scriptstyle\times10^{-3}}$ & $\bm{(4.5{\scriptstyle\pm0.4})}$&$\bm{{\scriptstyle\times10^{-4}}}$ \\
FPUT-Morse & $(6.2{\scriptstyle\pm0.4})$&${\scriptstyle\times10^{-3}}$ & \multicolumn{2}{c}{--} & $(7.1{\scriptstyle\pm1.2})$&${\scriptstyle\times10^{-3}}$ & $\bm{(2.3{\scriptstyle\pm0.6})}$&$\bm{{\scriptstyle\times10^{-4}}}$ \\
Fast-Slow & $(9.8{\scriptstyle\pm0.2})$&${\scriptstyle\times10^{-7}}$ & \multicolumn{2}{c}{--} & $(6.1{\scriptstyle\pm1.4})$&${\scriptstyle\times10^{-7}}$ & $\bm{(3.1{\scriptstyle\pm0.6})}$&$\bm{{\scriptstyle\times10^{-7}}}$ \\
\bottomrule
\end{tabular}

%% file: sections/6_conclusion.tex
\section{Conclusion}\label{sec:conclusion}

Hamiltonian structure determines how a learned energy function generates dynamics, but accurate long-term prediction ultimately depends on how well that energy itself is represented. We show that a monolithic neural representation of the Hamiltonian inherits the spectral bias of coordinate-based neural networks, and because the dynamics depend on gradients of the energy, its unresolved fine-scale components translate directly into substantial vector-field and rollout errors. LM-HNN addresses this by decomposing the Hamiltonian into kinetic and potential components represented by shared local energy functions over multiple interaction scales. As a result, LM-HNN extends the effective spectral range of the learned Hamiltonian, consistently improves derivative and long-horizon rollout accuracy across low-dimensional, interacting, and fast–slow systems, and remains accurate at unseen energy levels.

\textbf{Future Work.} LM-HNN’s shared local energy decomposition could be extended to more general interaction structures and systems described by spatially varying fields.

\textbf{Limitations.} The current formulation assumes conservative dynamics and does not explicitly model dissipation or external forcing. Additionally, the current formulation assumes access to the full system state and does not explicitly address partially observed dynamics.

%% file: sections/n0_statement.tex
\subsection*{AI use statement}
In this work, we used generative AI tools for generating synthetic datasets, implementing methods, supporting qualitative and thematic data analysis, and designing and running experiments. We have not used generative AI tools for developing theoretical models or conceptual frameworks, formulating mathematical claims, providing critical ingredients for proving mathematical claims, or interpreting results, and assisting in writing proofs, proposing or refining hypotheses, assisting with translation, cleaning and reformatting datasets are not applicable to this work. Additionally, we used generative AI tools for creating or editing software code, drafting parts of the manuscript, and editing the manuscript to improve readability. We have reviewed all AI-assisted work. The authors reviewed the implementation and experimental procedures, verified the reported results, and checked the manuscript for accuracy. We take responsibility for the final content of this work, including text, claims or artifacts produced with the aid of generative AI.

\subsection*{Ethics statement}
This work is methodological. All data are generated by numerically integrating analytic Hamiltonians of textbook mechanical systems, namely pendulums, nonlinear lattices, and a fast--slow oscillator, so the study involves no human subjects, personal data, or content that could identify individuals. The intended use is the data-driven modeling and simulation of conservative physical systems, and we are not aware of a direct path from the method to harmful applications. As with any learned dynamics model, the accuracy of LM-HNN is established only within the regimes evaluated here, and deployment in safety-relevant settings, such as the control of physical systems, would require validation on the target system. Our use of generative AI tools is disclosed in the AI use statement.

\subsection*{Reproducibility statement}
We describe the experimental settings and implementation details to support reproducibility, and will release the source code alongside the paper upon acceptance. Section~\ref{sec:method} specifies the model formulation, and Section~\ref{sec:exp-systems} introduces the benchmark systems and baselines. Appendix~\ref{app:systems} provides the Hamiltonian definitions, data-generation procedures, and training and test energy ranges. Appendix~\ref{app:training-protocol} describes the data splits, optimization and model-selection procedures, numerical integration, rollout horizons, evaluation metrics, and reporting across independent training seeds. Appendix~\ref{app:spectral-protocol} defines the spectral diagnostics and their computation, while Appendix~\ref{app:ablation} describes the ablation configurations and controlled comparisons. Model parameter counts are reported in Table~\ref{tab:systems-params}.

%% file: sections/n1_appendix.tex
\section{Systems, Data Generation, and Evaluation Protocol}
\label{app:data}

This appendix describes the experimental setup underlying the results in the main text, including how trajectories from the benchmark systems are generated and used for model training and evaluation. The spectral diagnostics complement these evaluations by examining errors in the learned Hamiltonians and their derivatives.

\subsection{Training and Evaluation Protocol}
\label{app:training-protocol}

\paragraph{Data generation.}
Unless stated otherwise, trajectories are generated with the classical fourth-order Runge--Kutta (RK4) method from initial conditions sampled on prescribed energy shells,
\[
H(\bz_0)=E.
\]
States from the generated training trajectories are pooled and randomly split into $80\%$ training and $20\%$ validation sets. For Hamiltonian models, analytic derivatives
\[
\dot{\bz}=J\nabla H(\bz)
\]
are used as supervision, and the models are trained with the derivative loss in Eq.~\eqref{eq:deriv-loss}. SympNet is trained on the same underlying trajectories using its one-step prediction objective, while FS-HNN follows its original multiscale training procedure on the same datasets.

\paragraph{Model inputs.} On a chain with fixed ends $q_0=q_{N+1}=0$, the potential sums over bonds $b=0,\dots,N$ and its local variables are the strain window $\mathcal S_b^r=(s_{b-r},\dots,s_{b+r})$ with $s_b=q_{b+1}-q_b$, zero-padded at the ends, so the neighborhood of Eq.~\eqref{eq:local-neighborhood} is made of differences of neighboring positions; the kinetic energy sums over sites and uses the position window $(q_{i-r},\dots,q_{i+r})$, joined by the momentum window in Eq.~\eqref{eq:kinetic-sum}. At $r=0$ the potential network sees one strain, so $v_{\theta,0}(s_b)$ is a nearest-neighbor bond potential, and the inverse mass of site $i$ depends on $q_i$ alone; the same $\mathcal R$ is used for both components. On the angle systems each angle enters every local function as $(\sin q,\cos q)$ and the momenta only through the quadratic form: $T_\theta=\tfrac12 m_\theta^{-1}(q)\,p^2$ for the pendulum, and $T_\theta=\tfrac12\bp^\top M_\theta^{-1}(q_1-q_2)\,\bp$ with a Cholesky-factorized $M_\theta^{-1}$ and $V_\theta=v_1(q_1)+v_2(q_2)+v_{12}(q_1-q_2)$ for the double pendulum. The HNN and its Fourier-feature and SIREN variants receive the same inputs as LM-HNN; SympNet and FS-HNN act on the raw state, with angle trajectories unwrapped so that they remain trajectories of the same system.

\paragraph{Optimization and model selection.}
Hamiltonian models are optimized with AdamW and a cosine learning-rate schedule. We train for 2000 epochs on the chain systems and for 600--1000 epochs on the remaining systems. Model selection is based on the validation derivative error. The derivative MSE reported in the tables is the squared error averaged over the state components as well as over the states, that is, the training loss of Eq.~\eqref{eq:deriv-loss} divided by the state dimension $2N$; the RMS vector-field error of the spectral diagnostics is the square root of $2N$ times this quantity. Unless stated otherwise, all reported results are the mean $\pm$ standard deviation over three independently trained model seeds.

\paragraph{Rollout evaluation.}
Generalization is evaluated from initial conditions that are not used in any training trajectory. In-distribution test initial conditions are sampled independently from the same energy ranges used for training, while the unseen-energy experiments use energy levels above the training range as specified in Appendix~\ref{app:systems}.

For Hamiltonian models, rollouts are generated by integrating the learned vector field with RK4. SympNet instead predicts the discrete evolution directly and is iterated autoregressively. We use 120 rollout steps for the chain systems, 200 steps for the pendulum systems, and 1000 steps for the fast--slow oscillator. For the fast--slow system, evaluation uses the 30 held-out reference test trajectories.

We report the final-step normalized rollout error,
\begin{equation}
    \mathcal E_{\mathrm{roll}}
    =
    \frac{
    \mathrm{MSE}\!\left(\hat{\bz}(T),\bz(T)\right)
    }{
    \operatorname{Var}\!\left(\bz(t)\right)_{t\in[0,T]}
    },
    \label{eq:normalized-rollout}
\end{equation}
where the denominator is the variance of the corresponding reference trajectory over the rollout horizon.

For the chain systems, we additionally measure errors in the predicted trajectory amplitudes using
\begin{equation}
    \mathcal E_{\mathrm{peak}}
    =
    \frac{1}{2}
    \frac{
    |\max_t \hat x(t)-\max_t x(t)|
    +
    |\min_t \hat x(t)-\min_t x(t)|
    }{
    \max_t x(t)-\min_t x(t)
    },
    \label{eq:peak-error}
\end{equation}
averaged over all state variables. We refer to this quantity as the \emph{peak error}.

The two measures answer different questions, and we report both because a model can do well on one while failing on the other. The normalized rollout error compares states at a single instant, so it is dominated by phase: once a predicted trajectory drifts in phase, its pointwise difference from the reference saturates at the scale of the motion itself, whether or not the predicted motion still has the right extent. The peak error instead compares the turning points of the motion and does not depend on when they are reached. This distinction is physically meaningful here, because in a conservative system the amplitude of a trajectory is fixed by the level set of the Hamiltonian passing through its initial state. A model whose predicted extrema are wrong has therefore learned an energy surface of the wrong shape, rather than a correct trajectory that has merely fallen out of step. On the chains the turning points are also where the anharmonic part of the bond potential is probed most strongly, so the peak error reports directly on the part of the potential that a smooth global approximation resolves least well. We report the peak error for the chain systems only: the angular coordinates of the pendulum systems are wrapped to $(-\pi,\pi]$, so their extrema carry no amplitude information.

\subsection{Dynamical Systems and Data Generation}
\label{app:systems}

\paragraph{Pendulum.}
We consider the unit-mass, unit-length pendulum with unit gravitational acceleration,
\begin{equation}
    H(q,p)
    =
    \frac{1}{2}p^2 + (1-\cos q).
\end{equation}
The training set contains sixteen energy shells evenly spaced over
$E\in[0.05,4.0]$, spanning both librational and rotational motion; the separatrix occurs at $E=2$.

For each energy shell, we generate eight initial conditions. The angle is sampled uniformly from $(-\pi,\pi]$ and rejected whenever $V(q)>E$. The momentum is then determined by the remaining kinetic energy,
\begin{equation}
    p
    =
    \pm\sqrt{2(E-V(q))},
\end{equation}
with the sign chosen randomly. Each initial condition is integrated for 300 steps with $\Delta t=0.05$, and the angle is wrapped to $(-\pi,\pi]$, yielding 38{,}400 states in total.

In-distribution rollouts start from two unseen initial conditions on each of the shells
\[
E\in\{0.5,1.5,2.5,3.5\},
\]
and run for 200 steps. Angular errors are computed modulo $2\pi$. For unseen-energy evaluation, we use
\[
E\in\{6,8\},
\]
both above the maximum training energy of $4$.

\paragraph{Double pendulum.}
We use a double pendulum with
$m_1=m_2=l_1=l_2=g=1$ and canonical coordinates
$(q_1,q_2,p_1,p_2)$. Its Hamiltonian is
\begin{align}
    H(\bq,\bp)
    &=
    \frac{1}{2}\bp^\top M(\bq)^{-1}\bp
    +(m_1+m_2)gl_1(1-\cos q_1)
    +m_2gl_2(1-\cos q_2),
    \\
    M(\bq)
    &=
    \begin{pmatrix}
    (m_1+m_2)l_1^2
    &
    m_2l_1l_2\cos(q_1-q_2)
    \\
    m_2l_1l_2\cos(q_1-q_2)
    &
    m_2l_2^2
    \end{pmatrix}.
\end{align}

Training uses sixteen energy shells evenly spaced over
$E\in[0.5,5.0]$, with eight initial conditions on each shell. The angles are sampled uniformly and rejected if $V(\bq)>E$. A random direction $\mathbf u$ in velocity space is then scaled so that
\begin{equation}
    \frac{1}{2}s^2\mathbf u^\top M(\bq)\mathbf u
    =
    E-V(\bq),
\end{equation}
after which the canonical momentum is set to
\begin{equation}
    \bp=M(\bq)\,s\mathbf u.
\end{equation}
Each trajectory contains 300 steps with $\Delta t=0.02$.

In-distribution rollouts use two unseen initial conditions from each of
\[
E\in\{1,2,3,4\}
\]
and run for 200 steps. The unseen-energy shells are
\[
E\in\{7,9\},
\]
compared with a maximum training energy of $5$.

\paragraph{FPUT, Toda, and Morse chains.}
We consider fixed-boundary chains of $N$ unit masses with
$q_0=q_{N+1}=0$. Defining the bond strain
\[
s_b=q_{b+1}-q_b,
\qquad
b=0,\ldots,N,
\]
the Hamiltonian takes the form
\begin{equation}
    H(\bq,\bp)
    =
    \frac{1}{2}\bp^\top A\bp
    +
    \sum_{b=0}^{N}\phi(s_b).
    \label{eq:chain-hamiltonian}
\end{equation}
We use $A=I$ except in the kinetic-coupling ablation of Table~\ref{tab:locality}.

For the FPUT chain,
\begin{equation}
    \phi_{\mathrm{FPUT}}(s)
    =
    \frac{1}{2}s^2
    +
    \frac{\alpha}{3}s^3
    +
    \frac{\beta}{4}s^4,
\end{equation}
with $(\alpha,\beta)=(0.25,0.10)$ unless stated otherwise. We additionally consider the Toda potential
\begin{equation}
    \phi_{\mathrm{Toda}}(s)
    =
    e^s-s-1,
\end{equation}
and the Morse potential
\begin{equation}
    \phi_{\mathrm{Morse}}(s)
    =
    D(1-e^{-as})^2,
    \qquad
    D=8,
    \qquad
    a=\frac{1}{\sqrt{2D}},
\end{equation}
where the parameters are chosen so that all three interaction potentials have unit curvature at $s=0$.

For the controlled spectral-bias experiment of Section~\ref{sec:exp-bias}, we augment the FPUT interaction with an oscillatory component,
\begin{equation}
    \phi_k(s)
    =
    \phi_{\mathrm{FPUT}}(s)
    +
    \frac{a}{k}\bigl(1-\cos ks\bigr),
    \qquad
    a=0.5.
    \label{eq:ripple}
\end{equation}
The corresponding bond force therefore contains the additional term
$a\sin(ks)$. Its amplitude is fixed as $k$ varies, while the energy amplitude of the perturbation decreases as $1/k$. This construction allows us to vary the wavenumber of the Hamiltonian perturbation without simultaneously increasing the magnitude of the associated force.

For the receptive-field study of Appendix~\ref{app:ablation} we also use a chain carrying two interaction ranges at once, obtained by adding a harmonic second-neighbor coupling to the FPUT bonds,
\begin{equation}
    H(\bq,\bp)
    =
    \frac{1}{2}\|\bp\|^2
    +
    \sum_{b=0}^{N}\phi_{\mathrm{FPUT}}(s_b)
    +
    \frac{g}{2}\sum_{i=0}^{N-1}\bigl(q_{i+2}-q_i\bigr)^2,
    \qquad
    g=0.3,
    \label{eq:twoscale}
\end{equation}
again with $q_0=q_{N+1}=0$. At the training amplitudes the second-neighbor term carries about $21\%$ of the potential energy, so it is a genuine but secondary contribution. The point of this system is that the two ranges are separated in the strain coordinates that the local potential heads read: since $q_{i+2}-q_i=s_i+s_{i+1}$, a window of radius $0$ contains a single strain and cannot represent the second-neighbor term at all, whereas radius $1$ and above can. Data generation and evaluation follow the protocol used for the other chains.

For each chain system, the training data contain eight energy shells evenly spaced over the energy-per-site range
\[
E/N\in[0.05,2.0].
\]
At each energy, we generate four trajectories of 100 steps with $\Delta t=0.05$.

Two trajectories begin from momentum-dominated states. Positions are sampled uniformly from
\[
\bq\in[-0.35,0.35]^N,
\]
and the remaining energy is assigned to the momentum along a random direction $\mathbf u$:
\begin{equation}
    \bp=s\mathbf u,
    \qquad
    \frac{1}{2}s^2\mathbf u^\top A\mathbf u
    =
    E-V(\bq).
\end{equation}

The other two trajectories begin from potential-dominated states. A random direction $\mathbf u$ in configuration space is scaled by bisection until
\begin{equation}
    V(s\mathbf u)=fE,
    \qquad
    f\sim\mathcal U(0.55,0.95),
\end{equation}
and the remaining energy is assigned to the momentum. These near-turning-point states increase coverage of the larger-strain configurations that determine the extrema of the resulting trajectories.

The data-generation seed is $100+N$. For $N=8$, the resulting training dataset contains 3{,}200 states. In-distribution rollouts start from two unseen momentum-dominated initial conditions at each of
\[
E/N\in\{0.25,0.75,1.25,1.75\}
\]
and run for 120 steps. Unseen-energy evaluation uses four initial conditions at each of
\[
E/N\in\{2.5,3.0\},
\]
both above the maximum training value of $2.0$.

\paragraph{Fast--slow oscillator.}
We use the nearly periodic Hamiltonian system of \citet{duruisseaux2023approximation}. With
$\bz=(q_1,q_2,p_1,p_2)$ and $\varepsilon=0.01$,
\begin{equation}
    H_\varepsilon
    =
    \frac{1}{2}\bigl(q_1^2+p_1^2\bigr)
    +
    \frac{\varepsilon}{2}\bigl(q_2^2+p_2^2\bigr)
    +
    \varepsilon q_1q_2\sin(2q_1+2q_2).
    \label{eq:fastslow-hamiltonian}
\end{equation}
The pair $(q_1,p_1)$ oscillates at unit frequency, while $(q_2,p_2)$ evolves on the slower scale $\varepsilon$, with nonlinear coupling of order $\varepsilon$.

We use the reference dataset of \citet{duruisseaux2023approximation}, consisting of 300 trajectories of 3000 steps with $\Delta t=0.1$. Initial conditions are drawn from three regions of phase space: one general window and two windows containing small slow-oscillator orbits centered to the left and right. The trajectories are divided into 240 training, 30 validation, and 30 test trajectories.

For derivative-trained Hamiltonian models, we subsample every twentieth state from the training and validation trajectories, yielding 40{,}770 states with analytic derivative targets. SympNet and FS-HNN use the full trajectories according to their respective training procedures.

Rollouts start from the initial states of the 30 held-out test trajectories and run for 1000 steps. In addition to the full-state error, we report the normalized rollout error of the slow pair $(q_2,p_2)$. The scaled-initial-condition test multiplies the same 30 initial states by $1.5$, which raises their energy by a factor between $2.1$ and $2.5$. The training trajectories have energies in $[0.04,2.48]$ and the test states in $[0.02,2.02]$; after scaling the energies lie in $[0.05,4.56]$, and 9 of the 30 states exceed the largest training energy. The other states move to larger amplitudes of both oscillators at energies that training still covers, so this row of Table~\ref{tab:generalization} measures extrapolation in amplitude rather than in energy alone.

\subsection{Spectral Diagnostics}
\label{app:spectral-protocol}

To characterize errors in the learned Hamiltonian directly, we evaluate several spectral diagnostics on 2048 validation states from each system using double precision.

Let
\begin{equation}
    \delta H(\bz)=H_\theta(\bz)-H(\bz)
\end{equation}
denote the Hamiltonian residual. Because the Hamiltonian is identifiable only up to an additive constant from derivative supervision, we subtract the mean of $\delta H$ before computing energy-error statistics. We report its RMS magnitude,
\begin{equation}
    \sigma_H
    =
    \sqrt{
    \mathbb E\!\left[
    \bigl(\delta H-\mathbb E[\delta H]\bigr)^2
    \right]
    },
\end{equation}
together with the RMS gradient error
\begin{equation}
    \|\nabla\delta H\|_{\mathrm{rms}}.
\end{equation}
Since the canonical symplectic matrix $J$ is orthogonal,
\begin{equation}
    \|J\nabla H_\theta-\dot{\bz}\|
    =
    \|J\nabla\delta H\|
    =
    \|\nabla\delta H\|,
\end{equation}
so this quantity is exactly the RMS vector-field error of the Hamiltonian model.

We additionally define the effective residual wavenumber
\begin{equation}
    k_{\mathrm{eff}}
    =
    \frac{
    \|\nabla\delta H\|_{\mathrm{rms}}
    }{
    \sigma_H
    }.
    \label{eq:keff}
\end{equation}
This quantity characterizes the characteristic spatial frequency of the remaining Hamiltonian residual: for fixed residual magnitude, larger $k_{\mathrm{eff}}$ corresponds to more rapidly varying error. It is therefore interpreted jointly with $\sigma_H$ and the vector-field error rather than as an accuracy metric on its own.

To inspect the residual spectrum more directly, we construct 96 one-dimensional probe lines through validation states. For chain systems, each line varies one site while holding the remaining coordinates fixed; for the pendulum systems, one angular coordinate is varied. Along each line, we compute the power spectra of the target Hamiltonian $H$, the residual $\delta H$, and the corresponding directional derivative of the residual using a Blackman--Harris window. We define $k_{99}$ as the wavenumber below which $99\%$ of the target Hamiltonian power is contained.

The heat maps in Figure~\ref{fig:spectral_bias_combined} are constructed as follows. For the perturbed FPUT chain, we quantify recovery of the added wavenumber-$k$ force component using
\begin{equation}
    \lambda
    =
    \frac{
    \mathrm{MSE}_0
    -
    \mathrm{MSE}_k
    +
    \mathrm{MSE}_{\mathrm{ignored}}
    }{
    2\,\mathrm{MSE}_{\mathrm{ignored}}
    },
    \label{eq:force-recovery}
\end{equation}
where $\mathrm{MSE}_k$ is the derivative error with respect to the full perturbed vector field, $\mathrm{MSE}_0$ is the error with respect to the perturbation-free vector field, and $\mathrm{MSE}_{\mathrm{ignored}}$ is the error incurred by omitting the oscillatory perturbation entirely. When the error in the smooth component is orthogonal to the perturbation, $\lambda$ equals the fraction of the perturbation force recovered by the learned model.

\section{Ablations}
\label{app:ablation}

We perform a sequence of ablations to identify which aspects of LM-HNN are responsible for its improvement and under what conditions the proposed representation is most effective. We first isolate the contributions of kinetic--potential decomposition, locality, and weight sharing. We then study how the receptive field should be chosen, how the kinetic representation should adapt to different coupling structures, and how the advantage of LM-HNN changes with system size and interaction complexity. Finally, we compare the proposed physical decomposition with generic architectural remedies for spectral bias. Throughout this appendix the tables report the derivative MSE on validation states, the normalized final-step rollout error of Eq.~\eqref{eq:normalized-rollout}, and the peak error of Eq.~\eqref{eq:peak-error}, which measures how accurately the amplitude of the predicted motion is preserved.

\textbf{Contributions of decomposition, locality, and weight sharing.}
We first ablate the three main structural components of LM-HNN on the FPUT and Toda chains. Table~\ref{tab:ablation} shows that the gain does not come from a single architectural modification, but from combining physically meaningful decomposition with local and shared representations.

\begin{table}[t]
    \caption{\textbf{Ablation of the LM-HNN structure.} We perform ablation on the FPUT and FPUT-Toda chains ($N=8$). Removing locality causes the largest degradation, followed by removing the kinetic--potential decomposition. Removing weight sharing increases the parameter count substantially without providing a consistent accuracy gain. Single-radius variants are included here for completeness and studied in more detail in Table~\ref{tab:radius}.}
    \label{tab:ablation}
    \setlength{\tabcolsep}{3pt}
    \vspace{2mm}
    \centering

\input{tables/table_ablation}
    \vspace{-5mm}
\end{table}

\begin{table}[b]
    \vspace{-5mm}
    \caption{\textbf{Ablation of the receptive-field.} We perform ablation on two chains with different interaction ranges ($N=8$, $\alpha=0.25$, $\beta=0.10$): the nearest-neighbor FPUT chain, and the two-scale chain of Eq.~\eqref{eq:twoscale}. The best single radius changes between the two systems, and the radius that wins on the nearest-neighbor chain, $\mathcal R=\{0\}$, is the one that cannot represent the second-neighbor interaction at all. The multiscale set $\mathcal R=\{0,1,2\}$ is close to the best choice on both chains and attains the lowest derivative error of any model on the two-scale chain.}
    \label{tab:radius}
    \vspace{2mm}
    \centering
    \input{tables/table_radius}
\end{table}

Removing the kinetic--potential split and replacing the separate energy branches with a single local network over joint position--momentum neighborhoods substantially increases both derivative and rollout errors. The degradation is larger when locality is removed entirely and the kinetic and potential energies are represented by global networks over the full state. These results indicate that kinetic--potential decomposition simplifies the learning problem, while locality provides the larger gain by reducing the dimensionality of the functions that each network must approximate.

Weight sharing plays a complementary role. Assigning independent networks to different sites and bonds increases the parameter count by nearly an order of magnitude without yielding a consistent improvement. Sharing the local interaction functions therefore acts not merely as a parameter-saving device, but as an appropriate inductive bias for systems built from repeated interactions.

\begin{table}[t]
    \caption{\textbf{Kinetic coupling and kinetic-head design for $N=8$.} Systems include $A=I$ for the standard FPUT chain; tridiagonal coupling $A=I+0.3\,\mathrm{tridiag}(1)$, which introduces $p_ip_{i+1}$ terms; and global coupling $A=I+\tfrac1N\mathbf 1\mathbf 1^\top$. We compare the diagonal quadratic form of Eq.~\eqref{eq:local-kinetic}, the local-summation form of Eq.~\eqref{eq:kinetic-sum}, and a dense Cholesky form $\tfrac12\bp^\top L_\theta(\bq)L_\theta(\bq)^\top\bp$. The results show that the kinetic representation should follow the coupling structure: diagonal for uncoupled momenta, local for finite-range coupling, and dense for global coupling.}
    \label{tab:locality}
    \setlength{\tabcolsep}{3pt}
    \vspace{2mm}
    \centering
    \input{tables/table_locality}
    \vspace{-5mm}
\end{table}

\textbf{Choice of receptive field.}
The previous ablation establishes the importance of locality, but a local model still needs a choice of interaction neighborhood. Table~\ref{tab:radius} compares single radii against the multiscale set on two chains whose interaction ranges differ: the nearest-neighbor FPUT chain, and the two-scale chain of Eq.~\eqref{eq:twoscale}, which adds a second-neighbor coupling to the same bonds.

On the nearest-neighbor chain, the radius matched to the interaction, $\mathcal R=\{0\}$, is the most accurate choice, and a single radius that is too large loses accuracy: $\mathcal R=\{2\}$ raises the rollout error by roughly two orders of magnitude. On the two-scale chain the ranking reverses. A radius-$0$ window holds one bond strain and therefore cannot represent the second-neighbor term at all; its derivative error stalls near $5\times10^{-2}$ across seeds and its rollout error exceeds that of the conventional HNN. The best single radius there is $\mathcal R=\{1\}$, the smallest window containing both interaction ranges.

The multiscale set is close to the best choice on both chains without being told which one it is. On the two-scale chain it reaches the lowest derivative error of any model in the table, below every single radius: the sharp nearest-neighbor potential is represented by the one-dimensional radius-$0$ branch while the smooth second-neighbor term is carried by the wider branches, so each component is learned in the coordinates it actually depends on. Its rollout error there stays within a factor of about two of the matched radius, and on the nearest-neighbor chain it remains an order of magnitude below the conventional HNN. Since the interaction range of a physical system is generally unknown in advance, and since the radius that wins on one chain is the one that fails on the other, the full model uses $\mathcal R=\{0,1,2\}$ throughout.

\textbf{Kinetic representation and coupling structure}
Locality is beneficial only when it reflects the structure of the underlying Hamiltonian. To test this explicitly, Table~\ref{tab:locality} varies the momentum coupling while changing the kinetic head accordingly. For the standard FPUT system with diagonal kinetic energy, the diagonal quadratic form in Eq.~\eqref{eq:local-kinetic} is both sufficient and most accurate. Once nearest-neighbor momentum coupling is introduced, a diagonal head can no longer represent the correct kinetic energy, while the local-summation form of Eq.~\eqref{eq:kinetic-sum} with $\mathcal R=\{0,1,2\}$ recovers the coupled dynamics accurately.

For genuinely global momentum coupling, however, the local-summation representation is no longer sufficient and a dense Cholesky parameterization performs substantially better. Thus, LM-HNN does not assume that every Hamiltonian is local. Rather, its advantage arises when the representation is matched to the interaction structure present in the system.

\textbf{Scaling with system size.}
A central benefit of the shared local representation is that its parameter count does not grow with the number of repeated interactions. We test whether this structural advantage also translates into better accuracy as the chain becomes longer. Table~\ref{tab:nscale} compares LM-HNN with parameter-matched conventional HNNs from $N=4$ to $N=16$.

The conventional HNN becomes progressively harder to train as the dimension of the global Hamiltonian increases: its derivative error rises by roughly an order of magnitude and its rollout error grows from approximately $3\times10^{-2}$ to $5\times10^{-1}$. In contrast, LM-HNN uses the same 26{,}892 parameters for every chain length and its rollout error remains in the $10^{-3}$ range throughout the sweep. The widening gap supports the interpretation that LM-HNN learns a reusable interaction law whose complexity is largely independent of how many times that interaction is repeated.

\begin{table}[htbp]
    \caption{\textbf{Scaling with chain length $N$ at matched parameter count.} We scale the chain length on the FPUT system with $\alpha=0.25$ and $\beta=0.10$. From $N=4$ to $N=16$, the rollout error of LM-HNN remains between approximately $9\times10^{-4}$ and $6\times10^{-3}$, while that of the conventional HNN grows from $3\times10^{-2}$ to $5\times10^{-1}$.}
    \label{tab:nscale}
    \vspace{2mm}
    \setlength{\tabcolsep}{3pt}
    \centering
    \input{tables/table_nscale}
    \vspace{-3mm}
\end{table}

\textbf{Sensitivity to interaction nonlinearity.}
We next study whether the advantage of the local representation changes as the interaction law becomes more difficult to approximate. Table~\ref{tab:elastic} varies the nonlinear coefficients $(\alpha,\beta)$ of the FPUT potential while keeping the model comparison otherwise fixed.

As the interaction becomes more nonlinear, the conventional HNN degrades substantially: its peak error increases from $4.9\%$ to as high as $9.7\%$, and its derivative and rollout errors increase correspondingly. LM-HNN remains much more stable, with peak errors between approximately $0.8\%$ and $1.2\%$ across the sweep. This result is consistent with the proposed representation mechanism. Stronger nonlinearities make the full global energy landscape increasingly difficult to approximate, whereas LM-HNN continues to learn the same low-dimensional bond-level interaction law.

\begin{table}[b]
    \vspace{-5mm}
    \caption{\textbf{Parameter matched comparison of LM-HNN and HNN.} FPUT chain ($N=8$) across elastic parameters $(\alpha,\beta)$, comparing the parameter-matched HNN and LM-HNN. As the interaction becomes more nonlinear, the conventional HNN degrades substantially, while LM-HNN remains comparatively stable; its peak error stays between approximately $0.8\%$ and $1.2\%$ across the sweep.}
    \label{tab:elastic}
    \vspace{2mm}
    \setlength{\tabcolsep}{2pt}
    \centering
    \resizebox{\linewidth}{!}{\input{tables/table_elastic}}
\end{table}

\textbf{Comparison with generic spectral-bias interventions.}
Because one motivation for LM-HNN is to improve the representation of fine-scale Hamiltonian structure, we finally compare it with generic methods designed to alleviate spectral bias in coordinate networks. Table~\ref{tab:interv} applies random Fourier features and sinusoidal activations to the conventional HNN at matched parameter count.

For this comparison the derivative error is measured on held-out trajectories, generated by the same energy-shell sampler with a data seed never used in training, so that it reports the accuracy of the learned vector field on new trajectories rather than along the training ones; the rollout error starts from unseen initial conditions as everywhere else.

Random Fourier features raise both errors on both systems. Sinusoidal activations fit the training trajectories closely, but the field they learn between those trajectories is inaccurate, so their derivative error on held-out trajectories is large and their rollouts from unseen initial conditions are more than an order of magnitude worse than those of the original HNN. LM-HNN has the lowest derivative and rollout error in every column of Table~\ref{tab:interv}, on the chain by a factor of about fifty over the next model.

This comparison indicates that increasing access to high-frequency functions is not sufficient by itself. For Hamiltonian learning, the fine-scale structure must also be organized in a way that is consistent with the physical decomposition and interaction structure of the energy so that improvements in local fitting translate into accurate dynamics away from the observed trajectories.

\textbf{Ablation summary.}
Taken together, the ablations support four conclusions. First, locality provides the largest accuracy gain by reducing the dimensionality of the Hamiltonian representation, while kinetic--potential decomposition further simplifies the functions that must be learned. Second, weight sharing makes this local representation scalable by allowing the same interaction law to be reused across repeated structures without increasing the parameter count. Third, the local representation is most effective when its receptive field and kinetic head match the coupling structure of the underlying Hamiltonian. Fourth, when the interaction range is not known in advance, combining multiple neighborhood scales provides a robust alternative to selecting a single radius, remaining competitive with the best matched single-scale choice across different interaction structures.

These experiments also distinguish LM-HNN from generic approaches that increase the spectral capacity of a coordinate network. Its advantage grows with both system size and interaction nonlinearity, and cannot be reproduced simply by adding Fourier features or sinusoidal activations. Together, the results support the central claim of this work: accurate Hamiltonian learning requires not only sufficient capacity to represent fine-scale variation, but also a representation that organizes that variation according to the physical structure of the underlying energy.

%% file: tables/table_ablation.tex
\begin{tabular}{lr*{3}{r@{}l}}
\toprule
\multicolumn{8}{c}{FPUT} \\
\cmidrule(lr){1-8}
Variant & Params & \multicolumn{2}{c}{Deriv.\ MSE} & \multicolumn{2}{c}{Rollout MSE} & \multicolumn{2}{c}{Peak err.} \\
\midrule
LM-HNN, $\mathcal R=\{0,1,2\}$ (full)
& 26,892
& $(5.6{\scriptstyle\pm0.9})$&${\scriptstyle\times10^{-4}}$
& $(4.9{\scriptstyle\pm0.6})$&${\scriptstyle\times10^{-3}}$
& $0.83$&${\scriptstyle\pm0.06\%}$ \\

single radius $\mathcal R=\{0\}$
& 8,708
& $\bm{(2.6{\scriptstyle\pm0.4})}$&$\bm{{\scriptstyle\times10^{-4}}}$
& $\bm{(2.5{\scriptstyle\pm0.8})}$&$\bm{{\scriptstyle\times10^{-4}}}$
& $\bm{0.17}$&$\bm{{\scriptstyle\pm0.04\%}}$ \\

single radius $\mathcal R=\{2\}$
& 9,220
& $(1.5{\scriptstyle\pm0.2})$&${\scriptstyle\times10^{-3}}$
& $(2.0{\scriptstyle\pm0.8})$&${\scriptstyle\times10^{-2}}$
& $1.65$&${\scriptstyle\pm0.30\%}$ \\

no kinetic--potential split
& 26,736
& $(2.1{\scriptstyle\pm0.2})$&${\scriptstyle\times10^{-3}}$
& $(3.8{\scriptstyle\pm0.6})$&${\scriptstyle\times10^{-2}}$
& $1.58$&${\scriptstyle\pm0.14\%}$ \\

no locality (global $T$, $V$ networks)
& 26,932
& $(4.0{\scriptstyle\pm0.1})$&${\scriptstyle\times10^{-3}}$
& $(5.5{\scriptstyle\pm0.5})$&${\scriptstyle\times10^{-2}}$
& $3.00$&${\scriptstyle\pm0.14\%}$ \\

no weight sharing
& 228,537
& $(4.0{\scriptstyle\pm0.2})$&${\scriptstyle\times10^{-4}}$
& $(3.5{\scriptstyle\pm0.8})$&${\scriptstyle\times10^{-3}}$
& $0.59$&${\scriptstyle\pm0.05\%}$ \\

\midrule
\multicolumn{8}{c}{FPUT-Toda} \\
\cmidrule(lr){1-8}
Variant & Params & \multicolumn{2}{c}{Deriv.\ MSE} & \multicolumn{2}{c}{Rollout MSE} & \multicolumn{2}{c}{Peak err.} \\
\midrule
LM-HNN, $\mathcal R=\{0,1,2\}$ (full)
& 26,892
& $(4.5{\scriptstyle\pm0.4})$&${\scriptstyle\times10^{-4}}$
& $\bm{(6.7{\scriptstyle\pm4.7})}$&$\bm{{\scriptstyle\times10^{-3}}}$
& $\bm{0.85}$&$\bm{{\scriptstyle\pm0.17\%}}$ \\

no kinetic--potential split
& 26,736
& $(3.2{\scriptstyle\pm0.4})$&${\scriptstyle\times10^{-3}}$
& $(7.1{\scriptstyle\pm1.2})$&${\scriptstyle\times10^{-2}}$
& $2.67$&${\scriptstyle\pm0.19\%}$ \\

no locality (global $T$, $V$ networks)
& 26,932
& $(1.6{\scriptstyle\pm0.1})$&${\scriptstyle\times10^{-2}}$
& $(3.8{\scriptstyle\pm1.0})$&${\scriptstyle\times10^{-1}}$
& $7.19$&${\scriptstyle\pm0.67\%}$ \\

no weight sharing
& 228,537
& $\bm{(4.3{\scriptstyle\pm0.8})}$&$\bm{{\scriptstyle\times10^{-4}}}$
& $(1.9{\scriptstyle\pm0.2})$&${\scriptstyle\times10^{-2}}$
& $0.99$&${\scriptstyle\pm0.07\%}$ \\
\bottomrule
\end{tabular}

%% file: tables/table_radius.tex

\begin{tabular}{llr*{3}{r@{}l}}
\toprule
System & Radii $\mathcal R$ & Params & \multicolumn{2}{c}{Deriv.\ MSE} & \multicolumn{2}{c}{Rollout MSE} & \multicolumn{2}{c}{Peak err.} \\
\midrule
Nearest-neighbor & $\{0\}$ & 8,708 & $\bm{(2.6{\scriptstyle\pm0.4})}$&$\bm{{\scriptstyle\times10^{-4}}}$ & $\bm{(2.5{\scriptstyle\pm0.7})}$&$\bm{{\scriptstyle\times10^{-4}}}$ & $\bm{0.17}$&$\bm{{\scriptstyle\pm0.03\%}}$ \\
chain & $\{1\}$ & 8,964 & $(1.0{\scriptstyle\pm0.2})$&${\scriptstyle\times10^{-3}}$ & $(4.1{\scriptstyle\pm1.8})$&${\scriptstyle\times10^{-3}}$ & $0.72$&${\scriptstyle\pm0.20\%}$ \\
 & $\{2\}$ & 9,220 & $(1.5{\scriptstyle\pm0.1})$&${\scriptstyle\times10^{-3}}$ & $(2.0{\scriptstyle\pm0.6})$&${\scriptstyle\times10^{-2}}$ & $1.65$&${\scriptstyle\pm0.24\%}$ \\
 & $\{0,1,2\}$ & 26,892 & $(5.6{\scriptstyle\pm0.7})$&${\scriptstyle\times10^{-4}}$ & $(4.9{\scriptstyle\pm0.5})$&${\scriptstyle\times10^{-3}}$ & $0.83$&${\scriptstyle\pm0.05\%}$ \\
 & HNN & 26,971 & $(6.0{\scriptstyle\pm0.1})$&${\scriptstyle\times10^{-3}}$ & $(2.0{\scriptstyle\pm0.1})$&${\scriptstyle\times10^{-1}}$ & $4.86$&${\scriptstyle\pm0.11\%}$ \\
\midrule
Two-scale & $\{0\}$ & 8,708 & $(5.0{\scriptstyle\pm0.0})$&${\scriptstyle\times10^{-2}}$ & $(8.7{\scriptstyle\pm0.0})$&${\scriptstyle\times10^{-1}}$ & $9.56$&${\scriptstyle\pm0.01\%}$ \\
chain & $\{1\}$ & 8,964 & $(6.3{\scriptstyle\pm0.6})$&${\scriptstyle\times10^{-4}}$ & $\bm{(9.1{\scriptstyle\pm2.2})}$&$\bm{{\scriptstyle\times10^{-4}}}$ & $\bm{0.43}$&$\bm{{\scriptstyle\pm0.04\%}}$ \\
 & $\{2\}$ & 9,220 & $(1.1{\scriptstyle\pm0.2})$&${\scriptstyle\times10^{-3}}$ & $(5.4{\scriptstyle\pm2.0})$&${\scriptstyle\times10^{-3}}$ & $0.93$&${\scriptstyle\pm0.21\%}$ \\
 & $\{0,1,2\}$ & 26,892 & $\bm{(4.4{\scriptstyle\pm0.3})}$&$\bm{{\scriptstyle\times10^{-4}}}$ & $(2.1{\scriptstyle\pm0.3})$&${\scriptstyle\times10^{-3}}$ & $0.54$&${\scriptstyle\pm0.03\%}$ \\
 & HNN & 26,971 & $(4.9{\scriptstyle\pm0.1})$&${\scriptstyle\times10^{-3}}$ & $(1.3{\scriptstyle\pm0.1})$&${\scriptstyle\times10^{-1}}$ & $4.20$&${\scriptstyle\pm0.10\%}$ \\
\bottomrule
\end{tabular}

%% file: tables/table_locality.tex

\begin{tabular}{ll*{3}{r@{}l}}
\toprule
Kinetic coupling & Kinetic head & \multicolumn{2}{c}{Deriv.\ MSE} & \multicolumn{2}{c}{Rollout MSE} & \multicolumn{2}{c}{Peak err.} \\
\midrule
Standard FPUT & quadratic, diagonal & $\bm{(5.6{\scriptstyle\pm0.9})}$&$\bm{{\scriptstyle\times10^{-4}}}$ & $\bm{(4.9{\scriptstyle\pm0.6})}$&$\bm{{\scriptstyle\times10^{-3}}}$ & $\bm{0.83}$&$\bm{{\scriptstyle\pm0.06\%}}$ \\
($A=I$) & local sum, $\mathcal R=\{0\}$ & $(6.7{\scriptstyle\pm0.8})$&${\scriptstyle\times10^{-4}}$ & $(1.0{\scriptstyle\pm0.2})$&${\scriptstyle\times10^{-2}}$ & $1.04$&${\scriptstyle\pm0.12\%}$ \\
 & local sum, $\mathcal R=\{0,1,2\}$ & $(6.5{\scriptstyle\pm0.9})$&${\scriptstyle\times10^{-4}}$ & $(1.1{\scriptstyle\pm0.3})$&${\scriptstyle\times10^{-2}}$ & $0.95$&${\scriptstyle\pm0.10\%}$ \\
\midrule
Tridiagonal & quadratic, diagonal & $(9.0{\scriptstyle\pm0.0})$&${\scriptstyle\times10^{-2}}$ & $(1.6{\scriptstyle\pm0.0})$&${\scriptstyle\times10^{0}}$ & $16.11$&${\scriptstyle\pm0.01\%}$ \\
coupling & local sum, $\mathcal R=\{0\}$ & $(9.0{\scriptstyle\pm0.0})$&${\scriptstyle\times10^{-2}}$ & $(1.6{\scriptstyle\pm0.0})$&${\scriptstyle\times10^{0}}$ & $16.23$&${\scriptstyle\pm0.04\%}$ \\
 & local sum, $\mathcal R=\{0,1,2\}$ & $\bm{(6.6{\scriptstyle\pm0.8})}$&$\bm{{\scriptstyle\times10^{-4}}}$ & $\bm{(1.5{\scriptstyle\pm0.5})}$&$\bm{{\scriptstyle\times10^{-2}}}$ & $\bm{1.42}$&$\bm{{\scriptstyle\pm0.26\%}}$ \\
\midrule
Global & local sum, $\mathcal R=\{0,1,2\}$ & $(8.1{\scriptstyle\pm0.3})$&${\scriptstyle\times10^{-3}}$ & $(1.7{\scriptstyle\pm0.1})$&${\scriptstyle\times10^{-1}}$ & $5.50$&${\scriptstyle\pm0.08\%}$ \\
coupling & dense Cholesky & $\bm{(9.7{\scriptstyle\pm1.3})}$&$\bm{{\scriptstyle\times10^{-4}}}$ & $\bm{(2.9{\scriptstyle\pm0.3})}$&$\bm{{\scriptstyle\times10^{-2}}}$ & $\bm{2.04}$&$\bm{{\scriptstyle\pm0.07\%}}$ \\
\bottomrule
\end{tabular}

%% file: tables/table_nscale.tex

\begin{tabular}{l@{\hspace{12pt}}cc*{4}{r@{}l}}
\toprule
& \multicolumn{2}{c}{Params} & \multicolumn{4}{c}{Derivative MSE} & \multicolumn{4}{c}{Rollout MSE} \\
\cmidrule(lr){2-3}\cmidrule(lr){4-7}\cmidrule(lr){8-11}
$N$ & HNN & LM-HNN & \multicolumn{2}{c}{HNN} & \multicolumn{2}{c}{LM-HNN} & \multicolumn{2}{c}{HNN} & \multicolumn{2}{c}{LM-HNN} \\
\midrule
4 & 27,031 & 26,892 & $(9.7{\scriptstyle\pm0.8})$&${\scriptstyle\times10^{-4}}$ & $\bm{(1.5{\scriptstyle\pm0.2})}$&$\bm{{\scriptstyle\times10^{-4}}}$ & $(3.2{\scriptstyle\pm0.5})$&${\scriptstyle\times10^{-2}}$ & $\bm{(9.4{\scriptstyle\pm0.6})}$&$\bm{{\scriptstyle\times10^{-4}}}$ \\
6 & 27,005 & 26,892 & $(3.3{\scriptstyle\pm0.1})$&${\scriptstyle\times10^{-3}}$ & $\bm{(5.0{\scriptstyle\pm0.4})}$&$\bm{{\scriptstyle\times10^{-4}}}$ & $(9.6{\scriptstyle\pm2.6})$&${\scriptstyle\times10^{-2}}$ & $\bm{(2.6{\scriptstyle\pm0.4})}$&$\bm{{\scriptstyle\times10^{-3}}}$ \\
8 & 26,971 & 26,892 & $(6.0{\scriptstyle\pm0.1})$&${\scriptstyle\times10^{-3}}$ & $\bm{(5.6{\scriptstyle\pm0.9})}$&$\bm{{\scriptstyle\times10^{-4}}}$ & $(2.0{\scriptstyle\pm0.1})$&${\scriptstyle\times10^{-1}}$ & $\bm{(4.9{\scriptstyle\pm0.6})}$&$\bm{{\scriptstyle\times10^{-3}}}$ \\
10 & 26,929 & 26,892 & $(8.8{\scriptstyle\pm1.1})$&${\scriptstyle\times10^{-3}}$ & $\bm{(4.3{\scriptstyle\pm0.8})}$&$\bm{{\scriptstyle\times10^{-4}}}$ & $(2.3{\scriptstyle\pm0.2})$&${\scriptstyle\times10^{-1}}$ & $\bm{(5.1{\scriptstyle\pm1.6})}$&$\bm{{\scriptstyle\times10^{-3}}}$ \\
12 & 26,879 & 26,892 & $(1.1{\scriptstyle\pm0.0})$&${\scriptstyle\times10^{-2}}$ & $\bm{(7.0{\scriptstyle\pm1.7})}$&$\bm{{\scriptstyle\times10^{-4}}}$ & $(5.2{\scriptstyle\pm0.4})$&${\scriptstyle\times10^{-1}}$ & $\bm{(4.6{\scriptstyle\pm1.1})}$&$\bm{{\scriptstyle\times10^{-3}}}$ \\
16 & 26,755 & 26,892 & $(1.7{\scriptstyle\pm0.1})$&${\scriptstyle\times10^{-2}}$ & $\bm{(7.6{\scriptstyle\pm1.6})}$&$\bm{{\scriptstyle\times10^{-4}}}$ & $(5.2{\scriptstyle\pm0.4})$&${\scriptstyle\times10^{-1}}$ & $\bm{(6.0{\scriptstyle\pm1.4})}$&$\bm{{\scriptstyle\times10^{-3}}}$ \\
\bottomrule
\end{tabular}

%% file: tables/table_elastic.tex

\begin{tabular}{@{}l*{6}{r@{}l}@{}}
\toprule
& \multicolumn{4}{c}{Deriv.\ MSE} & \multicolumn{4}{c}{Rollout MSE} & \multicolumn{4}{c}{Peak err.\ (\%)} \\
\cmidrule(lr){2-5}\cmidrule(lr){6-9}\cmidrule(lr){10-13}
$(\alpha,\beta)$ & \multicolumn{2}{c}{HNN} & \multicolumn{2}{c}{LM-HNN} & \multicolumn{2}{c}{HNN} & \multicolumn{2}{c}{LM-HNN} & \multicolumn{2}{c}{HNN} & \multicolumn{2}{c}{LM-HNN} \\
\midrule
(0.25, 0.10) & $(6.0{\scriptstyle\pm0.1})$&${\scriptstyle\times10^{-3}}$ & $\bm{(5.6{\scriptstyle\pm0.9})}$&$\bm{{\scriptstyle\times10^{-4}}}$ & $(2.0{\scriptstyle\pm0.1})$&${\scriptstyle\times10^{-1}}$ & $\bm{(4.9{\scriptstyle\pm0.6})}$&$\bm{{\scriptstyle\times10^{-3}}}$ & $4.86$&${\scriptstyle\pm0.13}$ & $\bm{0.83}$&$\bm{{\scriptstyle\pm0.06}}$ \\
(0.50, 0.10) & $(6.6{\scriptstyle\pm0.6})$&${\scriptstyle\times10^{-3}}$ & $\bm{(8.7{\scriptstyle\pm2.5})}$&$\bm{{\scriptstyle\times10^{-4}}}$ & $(3.7{\scriptstyle\pm0.7})$&${\scriptstyle\times10^{-1}}$ & $\bm{(1.8{\scriptstyle\pm1.0})}$&$\bm{{\scriptstyle\times10^{-2}}}$ & $6.03$&${\scriptstyle\pm0.54}$ & $\bm{0.96}$&$\bm{{\scriptstyle\pm0.10}}$ \\
(1.00, 0.30) & $(1.8{\scriptstyle\pm0.0})$&${\scriptstyle\times10^{-2}}$ & $\bm{(1.2{\scriptstyle\pm0.2})}$&$\bm{{\scriptstyle\times10^{-3}}}$ & $(8.2{\scriptstyle\pm0.5})$&${\scriptstyle\times10^{-1}}$ & $\bm{(2.2{\scriptstyle\pm0.8})}$&$\bm{{\scriptstyle\times10^{-2}}}$ & $9.68$&${\scriptstyle\pm0.88}$ & $\bm{1.24}$&$\bm{{\scriptstyle\pm0.12}}$ \\
(0.25, 0.30) & $(2.3{\scriptstyle\pm0.2})$&${\scriptstyle\times10^{-2}}$ & $\bm{(9.0{\scriptstyle\pm1.6})}$&$\bm{{\scriptstyle\times10^{-4}}}$ & $(2.9{\scriptstyle\pm0.5})$&${\scriptstyle\times10^{-1}}$ & $\bm{(6.1{\scriptstyle\pm1.2})}$&$\bm{{\scriptstyle\times10^{-3}}}$ & $5.88$&${\scriptstyle\pm0.35}$ & $\bm{0.93}$&$\bm{{\scriptstyle\pm0.06}}$ \\
(0.00, 0.50) & $(4.6{\scriptstyle\pm0.4})$&${\scriptstyle\times10^{-2}}$ & $\bm{(7.5{\scriptstyle\pm0.8})}$&$\bm{{\scriptstyle\times10^{-4}}}$ & $(4.6{\scriptstyle\pm0.4})$&${\scriptstyle\times10^{-1}}$ & $\bm{(3.9{\scriptstyle\pm1.2})}$&$\bm{{\scriptstyle\times10^{-3}}}$ & $8.29$&${\scriptstyle\pm0.38}$ & $\bm{0.76}$&$\bm{{\scriptstyle\pm0.11}}$ \\
\bottomrule
\end{tabular}

%% file: sections/n2_appendix_ot.tex
\section{Additional Results}
\label{app:more}
\begin{table}[b]
    \caption{\textbf{Spectral diagnostics on validation states.} RMS Hamiltonian residual $\sigma_H$, effective residual wavenumber $k_{\mathrm{eff}}$, and RMS vector-field error $\|\nabla\delta H\|_{\mathrm{rms}}$ for HNN, FS-HNN, and LM-HNN. LM-HNN has the smallest Hamiltonian residual and vector-field error on every system. Its remaining residual can have a larger effective wavenumber because the lower-frequency error has already been substantially reduced. The diagnostics follow the Fourier-domain analysis of spectral bias in \citet{rahaman2019spectral} and the frequency principle of \citet{xu2020frequency}, which examine the residual of a coordinate network by wavenumber over its input domain and across training; $k_{\mathrm{eff}}$ is defined in Equation~\ref{eq:keff} from Parseval's identity, and the link between the wavenumber content of the residual and the vector-field error follows the kernel analysis of \citet{wang2022when}.}
    \label{tab:bias-networks}
    \centering
    \vspace{3mm}

\input{tables/table_bias_networks}
\end{table}

\begin{figure}[b]
    \vspace{-3mm}
    \centering
    \includegraphics[width=\linewidth]{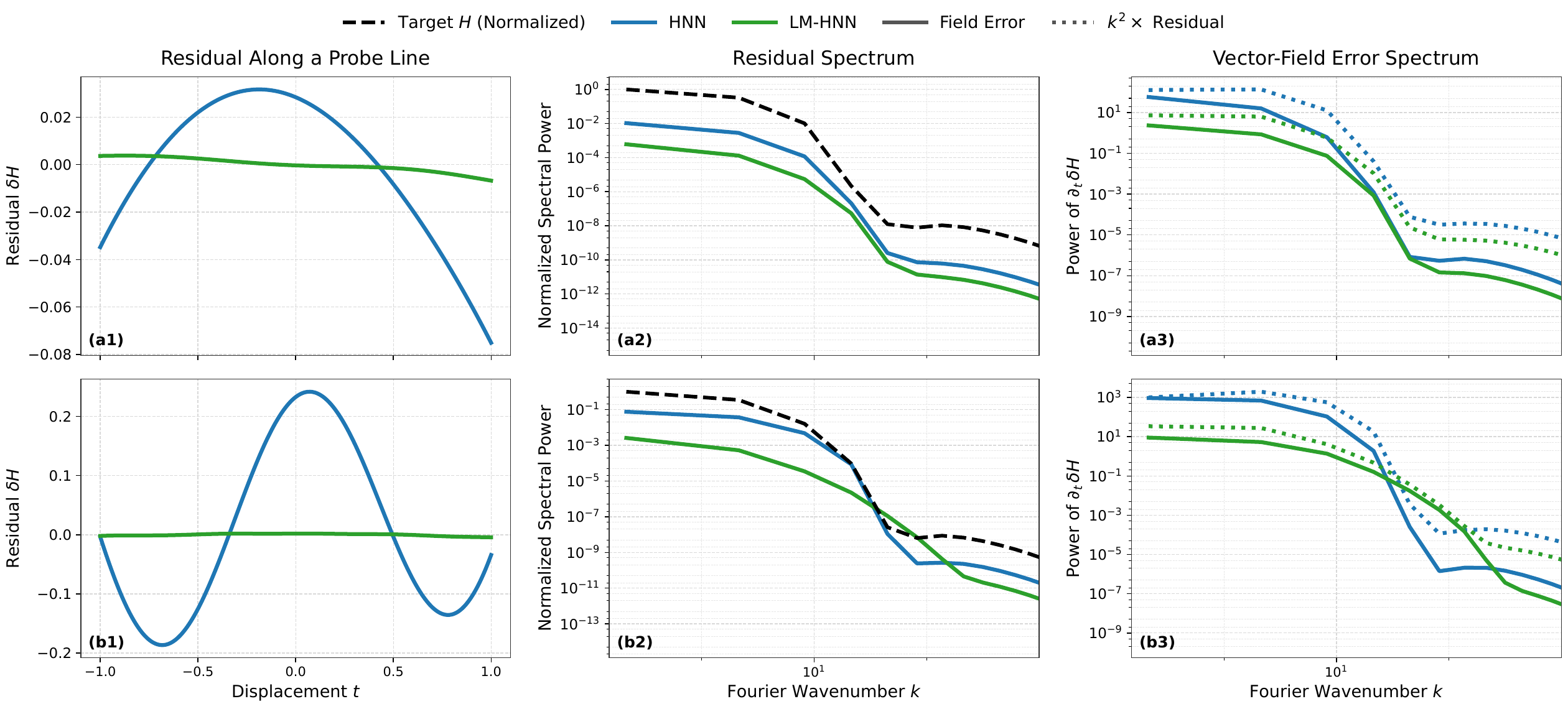}
    \caption{\textbf{Residual spectra along probe lines.} Left: the Hamiltonian residual $\delta H$ along a line obtained by displacing one site, for HNN and LM-HNN on the standard \textbf{(a)} FPUT chain and the \textbf{(b)} perturbed FPUT chain with $k=4$. Middle: power spectra of the target Hamiltonian and the residual of each model. Right: power spectrum of the corresponding vector-field error (solid) together with $k^2$ times the Hamiltonian-residual spectrum (dotted). Differentiation amplifies higher-wavenumber residuals, while LM-HNN reduces the residual over the spectral range occupied by the target.}
    \label{fig:spectra}
\end{figure}

This section provides additional evidence supporting the main experimental results. 
We first examine the Hamiltonian representation and its spectral error, then study the controlled perturbed-FPUT test, report accuracy across all systems, and finally relate Hamiltonian approximation error to rollout performance and generalization beyond the training energy range. We conclude with detailed results on the fast--slow oscillator and a preliminary extension to field-valued dynamics.

\subsection{Spectral Diagnostics of the Learned Hamiltonian}
\label{app:spectral-diagnostics}
Section~\ref{sec:exp-bias} tracks the residual spectrum of the learned Hamiltonian over training on the FPUT chain. This subsection applies the same view to all six systems at the end of training and connects the residual to the vector-field error that drives the rollouts. Table~\ref{tab:bias-networks} reports the RMS Hamiltonian residual $\sigma_{H}$, the effective residual wavenumber $k_{\mathrm{eff}}$ of Eq.~\eqref{eq:keff}, and the RMS vector-field error $\|\nabla\delta H\|_{\mathrm{rms}}$ on the validation states of Appendix~\ref{app:spectral-protocol} for HNN, FS-HNN, and LM-HNN. The diagnostics follow the Fourier-domain analysis of spectral bias in~\citet{rahaman2019spectral} and the frequency principle of~\citet{xu2020frequency}, which examine the residual of a coordinate network by wavenumber over its input domain, and the link between the wavenumber content of the residual and the vector-field error follows the kernel analysis of~\citet{wang2022when}. LM-HNN has the smallest Hamiltonian residual and the smallest vector-field error on every system, by more than an order of magnitude on the pendulum systems and by a factor of three to nine on the chains, with the narrowest margin on the fast--slow oscillator. The improvement observed in the rollouts is therefore already present in the learned energy and its gradient before any numerical integration is performed. FS-HNN, which decomposes the Hamiltonian by temporal frequency band, does not consistently reduce either quantity relative to the conventional HNN, which is consistent with the limitation lying in the phase-space wavenumber content of the energy rather than in the temporal spectrum of the trajectories.

The effective wavenumber behaves differently. LM-HNN matches or exceeds the $k_{\mathrm{eff}}$ of the conventional HNN on every system even though both its residual and its gradient error are smaller. This follows from the definition. Since $k_{\mathrm{eff}}$ is the ratio of the RMS gradient error to the RMS residual, removing the smooth part of the residual lowers the denominator by a larger fraction than the numerator and shifts the ratio upward, so the quantity reports where the remaining error is concentrated rather than how large it is. The residual that LM-HNN leaves behind sits at finer scales of the energy than the residual of the HNN, which is the expected outcome once the low-wavenumber modes that a coordinate network fits first have been resolved and only the slowest-learned components remain (Eq.~\eqref{eq:ntk-modes}). Because the gradient error of LM-HNN stays below that of the HNN on every system, this finer residual does not translate into a larger vector-field error, and $k_{\mathrm{eff}}$ is read jointly with $\sigma_{H}$ and $\|\nabla\delta H\|_{\mathrm{rms}}$ as Appendix~\ref{app:spectral-protocol} notes.

\begin{table}[b] 
    \vspace{-5mm}
    \caption{\textbf{Spectral-bias sweep on the perturbed FPUT chain ($N=8$).} We report the derivative and rollout errors of the HNN and LM-HNN against the perturbation wavenumber $k$, and the fraction of the perturbation force recovered, $1-\mathrm{MSE}/\mathrm{MSE}_{\mathrm{ignored}}$. The HNN loses the perturbation between $k=2$ and $k=4$; LM-HNN recovers it up to $k=6$.} 
    \label{tab:sweep} 
    \centering 
    \vspace{2mm}

\input{tables/table_sweep}
\end{table}
\begin{table}[t]
    \caption{\textbf{Rollout MSE on the perturbed FPUT ($N=8$) with increasing $k$.} LM-HNN achieves the lowest error at every tested wavenumber. At $k=4$, its rollout error remains below that of every competing baseline on the plain FPUT chain ($k=0$), demonstrating a substantially extended effective spectral range.}
    \label{tab:testbed}
    \centering
    \vspace{2mm}
    \input{tables/table_testbed}
    \vspace{-5mm}
\end{table}

Figure~\ref{fig:spectra} examines the residual directly along a representative probe line, obtained by displacing a single site of the chain while the remaining coordinates are held fixed, for the unperturbed FPUT chain and for the perturbed chain with $k=4$. On the unperturbed chain the residual of the HNN is smooth but far from zero along the whole line, while the residual of LM-HNN stays close to zero. The corresponding power spectra show that LM-HNN reduces the residual across the wavenumber band that carries the target Hamiltonian power rather than at one end of it. On the perturbed chain the residual of the HNN is itself an oscillation at the perturbation wavenumber, the signature of a component that the monolithic network has not resolved, and its residual spectrum concentrates its power near $k=4$. LM-HNN removes this component and its residual remains flat along the line. The right column relates these residuals to the dynamics. The power spectrum of the directional derivative of the residual coincides with $k^{2}$ times the residual spectrum, which confirms along the probe lines the relation of Eq.~\eqref{eq:parseval-grad} that underlies the effective wavenumber. Differentiation therefore weights the high-wavenumber part of the residual most heavily, and a residual that is small in energy but rapidly varying can dominate the vector-field error.

Taken together, the table and the probe-line spectra locate the advantage of LM-HNN at the level of the scalar Hamiltonian. The structured representation lowers the residual over the band occupied by the target, and because the vector field weights that band by $k^{2}$, the reduction appears in the gradient error before it appears in any rollout.
\subsection{Controlled Spectral-Bias Study on the Perturbed FPUT Chain}
\label{app:perturbed-fpu}

The diagnostics of the previous subsection measure the residual that each model leaves on a fixed Hamiltonian. The perturbed FPUT chain of Section~\ref{sec:exp-bias} instead varies the Hamiltonian itself, so that the wavenumber the model must resolve is prescribed rather than inferred. Adding the term $\frac{a}{k}\left(1-\cos ks\right)$ of Eq.~\eqref{eq:ripple} to the bond potential inserts a force component $a\sin\left(ks\right)$ whose amplitude is fixed at $a=0.5$ for every $k$, so the added structure carries the same dynamical weight at every wavenumber and differs only in how rapidly it varies with the bond strain. This subsection reports the quantities behind Figure~\ref{fig:spectral_bias_combined}(c,d), the derivative errors and recovery fractions of the two Hamiltonian models across the sweep in Table~\ref{tab:sweep}, and the rollout and peak errors of all four models in Tables~\ref{tab:testbed} and~\ref{tab:testbed-peak}.

\begin{table}[b]
    \vspace{-7mm}
    \caption{\textbf{Peak error on the perturbed FPUT ($N=8$) with variable perturbation wavenumber $k$.} LM-HNN achieves the lowest peak error at every tested wavenumber and remains below $1.2\%$ for $k\leq4$.}
    \label{tab:testbed-peak}
    \vspace{2mm}
    \centering
    \input{tables/table_testbed_peak}
\end{table}

Table~\ref{tab:sweep} lists the derivative error of each model against the perturbed vector field, the error incurred by omitting the perturbation entirely, and the fraction of the perturbation force recovered. The omission error is nearly constant across $k$, as expected for a force of fixed amplitude, so it provides a common reference for how much of the added component each model has learned. Both models recover the $k=1$ and $k=2$ components with only a small loss, and the heat maps of Figure~\ref{fig:spectral_bias_combined}(c,d) show that both learn the lowest wavenumbers first. The two diverge from $k=3$. The recovery of the conventional HNN falls steadily with $k$, and at $k=6$ its derivative error approaches the omission error, which means that within the training budget it has stopped resolving the component at all. LM-HNN recovers the perturbation nearly completely through $k=6$, with derivative errors that remain well below those of the HNN at every wavenumber, and its recovery declines only beyond that range. The ordering in which wavenumbers are learned is the same in both models, which is the ordering spectral bias predicts, and the local representation moves through it fast enough that the effective spectral limit shifts to a higher wavenumber within the same budget.

The same transition governs long-horizon prediction. Table~\ref{tab:testbed} reports the rollout error of all four models as $k$ increases, and LM-HNN has the lowest error at every tested wavenumber. Through $k=4$ its rollout error stays within a factor of three of its unperturbed value, which is below what the HNN and SympNet reach on the unperturbed chain. FS-HNN, which is accurate on the plain chain, loses an order of magnitude already at $k=2$. The peak error of Table~\ref{tab:testbed-peak} tells the same story for the amplitude of the motion. LM-HNN stays below $1.2\%$ through $k=4$, whereas the peak errors of the baselines rise to several percent and reach roughly $10\%$ at $k=4$ for the HNN and FS-HNN, so the extrema of the motion, which are fixed by the level set of the Hamiltonian through the initial state, are preserved only by the model that resolves the added structure of the potential. At $k=8$ and $k=16$ the perturbation lies beyond the effective spectral limit of every model, LM-HNN included, and all errors rise well above their levels at $k\le 4$, with LM-HNN remaining the lowest. The errors of all models are larger at $k=8$ than at $k=16$, which is consistent with the energy amplitude $2a/k$ of the perturbation shrinking with $k$, so that once the component is unresolved its net effect on the trajectory diminishes.

Across recovery fraction, derivative error, rollout error, and peak error, the sweep locates the same cutoff. The monolithic representation loses the added component between $k=2$ and $k=4$, while the local representation carries it to $k=6$ and degrades only beyond. The location of the cutoff depends on the training budget, as the heat maps show, but where it falls for a given budget is set by the representation, since both models share the data, the budget, and the parameter count.

\subsection{Accuracy Across Systems}
\label{app:accuracy}
\begin{figure}[t]
    \centering
    \includegraphics[width=\linewidth]{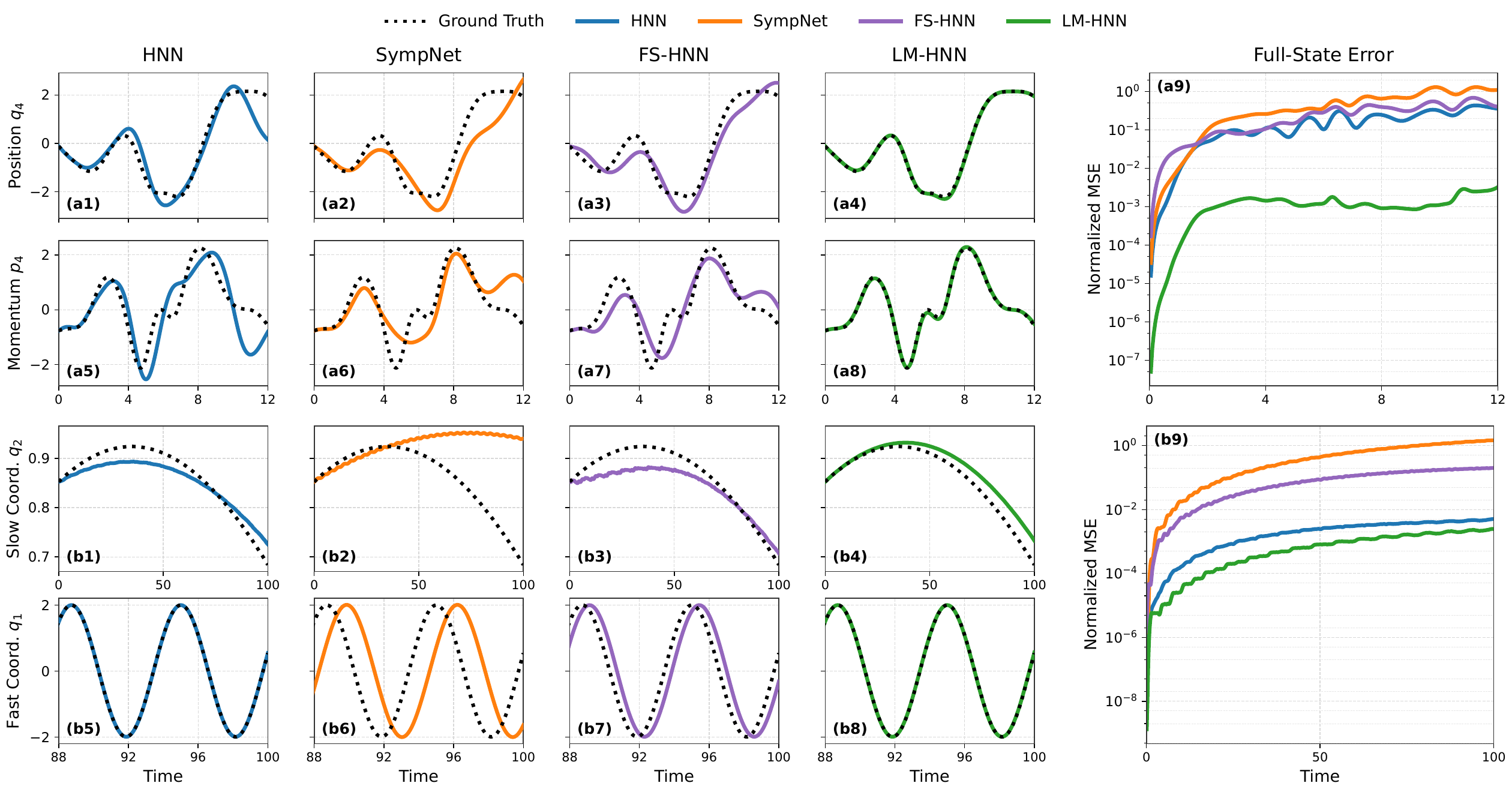}
    \vspace{-8mm}
    \caption{\textbf{Rollouts from unseen initial conditions.} \textbf{(a)} FPUT chain ($N=8$), one position coordinate, one momentum coordinate, and the normalized full-state error over 240 steps. \textbf{(b)} fast--slow oscillator ($\varepsilon=0.01$), the slow coordinate $q_2$ over the full horizon, the fast coordinate $q_1$ over the last two fast periods, and the normalized full-state error over 1000 steps. LM-HNN stays on the reference trajectory of the chain while the baselines accumulate phase and amplitude errors; on the fast--slow oscillator it has the lowest error over the whole horizon, following the slow drift most closely and keeping the fast phase, whereas the HNN drifts below the slow orbit in the middle of the horizon and SympNet and FS-HNN lose the fast phase.}
    \label{fig:rollout-case}
    \vspace{-3mm}
\end{figure}

We supplement the cross-system comparison in Section~\ref{sec:exp-accuracy} with model sizes and representative rollouts from unseen initial conditions. Table~\ref{tab:systems-params} reports the parameter counts of HNN, SympNet, FS-HNN, and LM-HNN across the six benchmark systems. The models use comparable parameter budgets within each system, showing that LM-HNN's improvements in derivative and rollout accuracy are achieved with a similar number of trainable parameters.

Figure~\ref{fig:rollout-case} (top) illustrates the evolution of prediction errors on the FPUT chain with $N=8$. The trajectory spans 240 steps, extending beyond the 120-step horizon used in the quantitative comparison. LM-HNN closely follows the reference position and momentum coordinates, while the baselines develop visible phase and amplitude errors. Its normalized full-state error also remains substantially lower over the illustrated horizon. These trajectories show that the improvement extends to both the timing and extent of the predicted motion.

Figure~\ref{fig:rollout-case} (bottom) examines the fast--slow oscillator with $\varepsilon=0.01$ over 1000 steps. The slow coordinate $q_2$ shows the gradual evolution of the orbit, while the final two periods of $q_1$ reveal accumulated phase differences in the fast component. LM-HNN captures the main slow variation and remains closely aligned with the reference fast oscillation. HNN deviates from the slow orbit around the middle of the horizon, while SympNet and FS-HNN exhibit visible phase shifts in the fast component. The corresponding full-state error remains lower for LM-HNN over the rollout.

\begin{table}[b]
    \vspace{-5mm}
    \caption{\textbf{Parameter counts of LM-HNN and baselines.} The models are chosen to use comparable parameter budgets so that the accuracy comparison primarily reflects differences in representation rather than raw model capacity.}
    \label{tab:systems-params}
    \vspace{2mm}
    \centering
    \input{tables/table_systems_params}
\end{table}

\begin{table}[t]
    \caption{\textbf{Hamiltonian and spectral diagnostics and rollout errors from unseen initial conditions inside and above the training energy range.} LM-HNN has smaller Hamiltonian residual, vector-field error, and rollout error across the tested systems. On several systems, much of its remaining residual power lies beyond the $k_{99}$ bandwidth containing $99\%$ of the target Hamiltonian power.}
    \label{tab:diag}
    \vspace{2mm}
    \setlength{\tabcolsep}{4.5pt}
    \centering

\input{tables/table_diag}
    \vspace{-5mm}
\end{table}
\subsection{Hamiltonian Error and Rollout Generalization}
\label{app:error-generalization}

We examine how the unseen-energy results in Section~\ref{sec:exp-generalization} relate to the accuracy of the learned Hamiltonian within the training range. Table~\ref{tab:diag} combines Hamiltonian and spectral diagnostics on validation states with normalized rollout errors from unseen initial conditions inside and above the training-energy range. Following Appendix~\ref{app:spectral-protocol}, the energy residual $\delta H=H_\theta-H$ is mean-centered before computing $\sigma_H$, removing the additive offset that is undetermined by derivative supervision.

Across the reported systems, LM-HNN achieves a smaller Hamiltonian residual $\sigma_H$ and lower RMS vector-field error $\|\nabla\delta H\|_{\mathrm{rms}}$ than HNN, accompanied by lower rollout errors in both evaluation regimes. The accuracy improvement is therefore already visible in the learned energy and its derivatives before numerical integration. The association with lower rollout error above the training-energy range further connects the quality of the learned Hamiltonian to the generalization behavior observed in the main experiments.

\begin{table}[t]
    \caption{\textbf{Fast--slow oscillator ($\varepsilon=0.01$) rollout error.} We report the rollout MSE over the 30 unseen test initial conditions after 200 and 1000 steps with $\Delta t=0.1$, reported for the full state and for the slow pair $(q_2,p_2)$. LM-HNN achieves the lowest error in every column, improving both the complete dynamics and the weak slow component.}
    \label{tab:fastslow}
    \setlength{\tabcolsep}{3pt}
    \vspace{2mm}
    \centering
    \input{tables/table_fastslow}
\end{table}

The spectral diagnostics characterize the variation remaining in these residuals. On the pendulum systems and the perturbed FPUT chain, most of LM-HNN's residual power lies beyond $k_{99}$, the bandwidth containing $99\%$ of the target Hamiltonian power. Together with the smaller residual magnitude, this is consistent with more accurate recovery of the dominant Hamiltonian structure. LM-HNN also has a larger effective residual wavenumber $k_{\mathrm{eff}}$ on several systems. As defined in Eq.~\eqref{eq:keff}, this quantity measures gradient error relative to residual magnitude, so a larger value can coexist with lower absolute energy and vector-field errors. These spectral quantities therefore describe the composition of the remaining error and should be interpreted alongside $\sigma_H$ and $\|\nabla\delta H\|_{\mathrm{rms}}$.

\subsection{Detailed Results on the Fast--Slow Oscillator}
\label{app:fastslow}

We examine prediction accuracy on the fast--slow system of Eq.~\eqref{eq:fastslow-hamiltonian}, with $\varepsilon=0.01$, by evaluating the full state and the slow pair $(q_2,p_2)$ separately. Table~\ref{tab:fastslow} reports normalized rollout errors after 200 and 1000 steps with $\Delta t=0.1$, using the initial conditions of the 30 held-out test trajectories. The separate slow-pair metric measures how accurately the gradual evolution of these variables is recovered, complementing the full-state metric, which also includes the fast oscillations.

LM-HNN achieves the lowest mean error for both the full state and the slow pair at both horizons. The improvement in full-state prediction is therefore accompanied by more accurate slow dynamics. Although errors increase as the rollout extends from 200 to 1000 steps, LM-HNN retains its advantage in both metrics. This complements the representative trajectories in Figure~\ref{fig:rollout-case}, where LM-HNN captures the main slow evolution while remaining closely aligned with the reference fast oscillation.

The comparison with FS-HNN is informative because that baseline explicitly separates trajectory information across temporal frequency bands. LM-HNN obtains lower mean errors for both state groups at both horizons, supporting the effectiveness of organizing the Hamiltonian around its physical interaction structure even when the resulting dynamics evolve on widely separated temporal scales.

%% file: tables/table_bias_networks.tex
\begin{tabular}{llr@{}lcr@{}l}
\toprule
System & Model & \multicolumn{2}{c}{$\sigma_H$} & $k_{\mathrm{eff}}$ & \multicolumn{2}{c}{$\|\nabla\delta H\|_{\mathrm{rms}}$} \\
\midrule
\multirow{3}{*}{Pendulum} & HNN & $(4.5{\scriptstyle\pm0.8})$&${\scriptstyle\times10^{-4}}$ & $4.1$ & $(1.9{\scriptstyle\pm0.4})$&${\scriptstyle\times10^{-3}}$ \\
 & FS-HNN & $(1.3{\scriptstyle\pm0.9})$&${\scriptstyle\times10^{-3}}$ & $2.9$ & $(3.0{\scriptstyle\pm0.6})$&${\scriptstyle\times10^{-3}}$ \\
 & LM-HNN & $\bm{(2.7{\scriptstyle\pm0.9})}$&$\bm{{\scriptstyle\times10^{-5}}}$ & $\bm{4.2}$ & $\bm{(1.1{\scriptstyle\pm0.5})}$&$\bm{{\scriptstyle\times10^{-4}}}$ \\
\midrule
\multirow{3}{*}{Double Pendulum} & HNN & $(1.5{\scriptstyle\pm0.2})$&${\scriptstyle\times10^{-2}}$ & $4.8$ & $(7.1{\scriptstyle\pm0.7})$&${\scriptstyle\times10^{-2}}$ \\
 & FS-HNN & $(8.4{\scriptstyle\pm1.5})$&${\scriptstyle\times10^{-2}}$ & $3.6$ & $(2.9{\scriptstyle\pm0.1})$&${\scriptstyle\times10^{-1}}$ \\
 & LM-HNN & $\bm{(3.8{\scriptstyle\pm1.5})}$&$\bm{{\scriptstyle\times10^{-4}}}$ & $\bm{9.1}$ & $\bm{(3.4{\scriptstyle\pm1.2})}$&$\bm{{\scriptstyle\times10^{-3}}}$ \\
\midrule
\multirow{3}{*}{FPUT} & HNN & $(1.7{\scriptstyle\pm0.1})$&${\scriptstyle\times10^{-1}}$ & $1.8$ & $(3.1{\scriptstyle\pm0.0})$&${\scriptstyle\times10^{-1}}$ \\
 & FS-HNN & $(1.7{\scriptstyle\pm0.7})$&${\scriptstyle\times10^{-1}}$ & $1.3$ & $(2.2{\scriptstyle\pm1.1})$&${\scriptstyle\times10^{-1}}$ \\
 & LM-HNN & $\bm{(4.4{\scriptstyle\pm0.0})}$&$\bm{{\scriptstyle\times10^{-2}}}$ & $\bm{2.1}$ & $\bm{(9.4{\scriptstyle\pm0.7})}$&$\bm{{\scriptstyle\times10^{-2}}}$ \\
\midrule
\multirow{3}{*}{FPUT-Toda} & HNN & $(2.7{\scriptstyle\pm0.1})$&${\scriptstyle\times10^{-1}}$ & $1.8$ & $(4.9{\scriptstyle\pm0.1})$&${\scriptstyle\times10^{-1}}$ \\
 & FS-HNN & $(1.5{\scriptstyle\pm0.1})$&${\scriptstyle\times10^{-1}}$ & $1.4$ & $(2.1{\scriptstyle\pm0.0})$&${\scriptstyle\times10^{-1}}$ \\
 & LM-HNN & $\bm{(3.0{\scriptstyle\pm0.4})}$&$\bm{{\scriptstyle\times10^{-2}}}$ & $\bm{2.9}$ & $\bm{(8.4{\scriptstyle\pm0.4})}$&$\bm{{\scriptstyle\times10^{-2}}}$ \\
\midrule
\multirow{3}{*}{FPUT-Morse} & HNN & $(2.7{\scriptstyle\pm0.3})$&${\scriptstyle\times10^{-1}}$ & $1.2$ & $(3.2{\scriptstyle\pm0.1})$&${\scriptstyle\times10^{-1}}$ \\
 & FS-HNN & $(2.1{\scriptstyle\pm0.3})$&${\scriptstyle\times10^{-1}}$ & $1.6$ & $(3.4{\scriptstyle\pm0.3})$&${\scriptstyle\times10^{-1}}$ \\
 & LM-HNN & $\bm{(3.2{\scriptstyle\pm0.5})}$&$\bm{{\scriptstyle\times10^{-2}}}$ & $\bm{1.9}$ & $\bm{(6.1{\scriptstyle\pm0.8})}$&$\bm{{\scriptstyle\times10^{-2}}}$ \\
\midrule
\multirow{3}{*}{Fast--slow} & HNN & $(2.7{\scriptstyle\pm0.1})$&${\scriptstyle\times10^{-4}}$ & $5.9$ & $(1.6{\scriptstyle\pm0.0})$&${\scriptstyle\times10^{-3}}$ \\
 & FS-HNN & $(1.2{\scriptstyle\pm0.1})$&${\scriptstyle\times10^{-3}}$ & $5.2$ & $(6.4{\scriptstyle\pm0.4})$&${\scriptstyle\times10^{-3}}$ \\
 & LM-HNN & $\bm{(1.7{\scriptstyle\pm0.2})}$&$\bm{{\scriptstyle\times10^{-4}}}$ & $\bm{6.9}$ & $\bm{(1.1{\scriptstyle\pm0.1})}$&$\bm{{\scriptstyle\times10^{-3}}}$ \\
\bottomrule
\end{tabular}

%% file: tables/table_sweep.tex
\begin{tabular}{l*{3}{r@{}l}}
\toprule
Perturbation wavenumber $k$ & \multicolumn{2}{c}{0} & \multicolumn{2}{c}{1} & \multicolumn{2}{c}{2} \\
\midrule
HNN derivative MSE
& $(6.0{\scriptstyle\pm0.1})$&${\scriptstyle\times10^{-3}}$
& $(3.3{\scriptstyle\pm0.2})$&${\scriptstyle\times10^{-3}}$
& $(1.1{\scriptstyle\pm0.1})$&${\scriptstyle\times10^{-2}}$ \\

LM-HNN derivative MSE
& $\bm{(5.6{\scriptstyle\pm0.9})}$&$\bm{{\scriptstyle\times10^{-4}}}$
& $\bm{(4.9{\scriptstyle\pm1.2})}$&$\bm{{\scriptstyle\times10^{-4}}}$
& $\bm{(2.1{\scriptstyle\pm0.5})}$&$\bm{{\scriptstyle\times10^{-3}}}$ \\

HNN rollout MSE
& $(2.0{\scriptstyle\pm0.1})$&${\scriptstyle\times10^{-1}}$
& $(1.2{\scriptstyle\pm0.1})$&${\scriptstyle\times10^{-1}}$
& $(2.0{\scriptstyle\pm0.2})$&${\scriptstyle\times10^{-1}}$ \\

LM-HNN rollout MSE
& $\bm{(4.9{\scriptstyle\pm0.6})}$&$\bm{{\scriptstyle\times10^{-3}}}$
& $\bm{(5.0{\scriptstyle\pm1.1})}$&$\bm{{\scriptstyle\times10^{-3}}}$
& $\bm{(1.2{\scriptstyle\pm0.3})}$&$\bm{{\scriptstyle\times10^{-2}}}$ \\

\midrule
Perturbation ignored (deriv. MSE)
& \multicolumn{2}{c}{--}
& $1.1$&${\scriptstyle\times10^{-1}}$
& $1.2$&${\scriptstyle\times10^{-1}}$ \\

HNN force recovered
& \multicolumn{2}{c}{--}
& \multicolumn{2}{c}{$97\%$}
& \multicolumn{2}{c}{$91\%$} \\

LM-HNN force recovered
& \multicolumn{2}{c}{--}
& \multicolumn{2}{c}{$\bm{100\%}$}
& \multicolumn{2}{c}{$\bm{98\%}$} \\

\midrule
Perturbation wavenumber $k$ & \multicolumn{2}{c}{3} & \multicolumn{2}{c}{4} & \multicolumn{2}{c}{6} \\
\midrule
HNN derivative MSE
& $(4.5{\scriptstyle\pm0.1})$&${\scriptstyle\times10^{-2}}$
& $(8.6{\scriptstyle\pm0.1})$&${\scriptstyle\times10^{-2}}$
& $(1.1{\scriptstyle\pm0.0})$&${\scriptstyle\times10^{-1}}$ \\

LM-HNN derivative MSE
& $\bm{(1.9{\scriptstyle\pm0.2})}$&$\bm{{\scriptstyle\times10^{-4}}}$
& $\bm{(8.6{\scriptstyle\pm2.9})}$&$\bm{{\scriptstyle\times10^{-4}}}$
& $\bm{(3.6{\scriptstyle\pm1.0})}$&$\bm{{\scriptstyle\times10^{-3}}}$ \\

HNN rollout MSE
& $(6.8{\scriptstyle\pm0.8})$&${\scriptstyle\times10^{-1}}$
& $(1.0{\scriptstyle\pm0.0})$&${\scriptstyle\times10^{0}}$
& $(1.3{\scriptstyle\pm0.1})$&${\scriptstyle\times10^{0}}$ \\

LM-HNN rollout MSE
& $\bm{(2.4{\scriptstyle\pm0.6})}$&$\bm{{\scriptstyle\times10^{-3}}}$
& $\bm{(1.4{\scriptstyle\pm0.7})}$&$\bm{{\scriptstyle\times10^{-2}}}$
& $\bm{(1.8{\scriptstyle\pm0.3})}$&$\bm{{\scriptstyle\times10^{-1}}}$ \\

\midrule
Perturbation ignored (deriv. MSE)
& $1.2$&${\scriptstyle\times10^{-1}}$
& $1.2$&${\scriptstyle\times10^{-1}}$
& $1.2$&${\scriptstyle\times10^{-1}}$ \\

HNN force recovered
& \multicolumn{2}{c}{$62\%$}
& \multicolumn{2}{c}{$26\%$}
& \multicolumn{2}{c}{$10\%$} \\

LM-HNN force recovered
& \multicolumn{2}{c}{$\bm{100\%}$}
& \multicolumn{2}{c}{$\bm{99\%}$}
& \multicolumn{2}{c}{$\bm{97\%}$} \\
\bottomrule
\end{tabular}

%% file: tables/table_testbed.tex

\begin{tabular}{l*{4}{r@{}l}}
\toprule
$k$ & \multicolumn{2}{c}{HNN} & \multicolumn{2}{c}{SympNet} & \multicolumn{2}{c}{FS-HNN} & \multicolumn{2}{c}{LM-HNN (ours)} \\
\midrule
0 & $(2.0{\scriptstyle\pm0.1})$&${\scriptstyle\times10^{-1}}$ & $(6.2{\scriptstyle\pm0.6})$&${\scriptstyle\times10^{-2}}$ & $(1.2{\scriptstyle\pm0.4})$&${\scriptstyle\times10^{-2}}$ & $\bm{(4.9{\scriptstyle\pm0.6})}$&$\bm{{\scriptstyle\times10^{-3}}}$ \\
$2$ & $(2.0{\scriptstyle\pm0.2})$&${\scriptstyle\times10^{-1}}$ & $(6.3{\scriptstyle\pm1.0})$&${\scriptstyle\times10^{-2}}$ & $(1.7{\scriptstyle\pm0.4})$&${\scriptstyle\times10^{-1}}$ & $\bm{(1.2{\scriptstyle\pm0.3})}$&$\bm{{\scriptstyle\times10^{-2}}}$ \\
$4$ & $(1.0{\scriptstyle\pm0.0})$&${\scriptstyle\times10^{0}}$ & $(1.2{\scriptstyle\pm0.4})$&${\scriptstyle\times10^{-1}}$ & $(6.7{\scriptstyle\pm2.5})$&${\scriptstyle\times10^{-1}}$ & $\bm{(1.4{\scriptstyle\pm0.7})}$&$\bm{{\scriptstyle\times10^{-2}}}$ \\
$8$ & $(8.5{\scriptstyle\pm0.4})$&${\scriptstyle\times10^{-1}}$ & $(8.3{\scriptstyle\pm0.2})$&${\scriptstyle\times10^{-1}}$ & $(9.8{\scriptstyle\pm0.3})$&${\scriptstyle\times10^{-1}}$ & $\bm{(5.0{\scriptstyle\pm0.1})}$&$\bm{{\scriptstyle\times10^{-1}}}$ \\
$16$ & $(4.2{\scriptstyle\pm0.2})$&${\scriptstyle\times10^{-1}}$ & $(3.4{\scriptstyle\pm0.5})$&${\scriptstyle\times10^{-1}}$ & $(4.9{\scriptstyle\pm0.4})$&${\scriptstyle\times10^{-1}}$ & $\bm{(1.8{\scriptstyle\pm0.0})}$&$\bm{{\scriptstyle\times10^{-1}}}$ \\
\bottomrule
\end{tabular}

%% file: tables/table_testbed_peak.tex

\begin{tabular}{l*{4}{r@{}l}}
\toprule
$k$ & \multicolumn{2}{c}{HNN} & \multicolumn{2}{c}{SympNet} & \multicolumn{2}{c}{FS-HNN} & \multicolumn{2}{c}{LM-HNN} \\
\midrule
0 & $4.86$&${\scriptstyle\pm0.13\%}$ & $2.12$&${\scriptstyle\pm0.11\%}$ & $2.99$&${\scriptstyle\pm0.85\%}$ & $\bm{0.83}$&$\bm{{\scriptstyle\pm0.06\%}}$ \\
$2$ & $5.16$&${\scriptstyle\pm0.56\%}$ & $1.55$&${\scriptstyle\pm0.10\%}$ & $4.08$&${\scriptstyle\pm1.18\%}$ & $\bm{1.16}$&$\bm{{\scriptstyle\pm0.12\%}}$ \\
$4$ & $11.37$&${\scriptstyle\pm0.27\%}$ & $3.16$&${\scriptstyle\pm0.87\%}$ & $9.22$&${\scriptstyle\pm1.19\%}$ & $\bm{0.69}$&$\bm{{\scriptstyle\pm0.20\%}}$ \\
$8$ & $9.99$&${\scriptstyle\pm0.53\%}$ & $9.91$&${\scriptstyle\pm0.83\%}$ & $10.82$&${\scriptstyle\pm0.37\%}$ & $\bm{7.16}$&$\bm{{\scriptstyle\pm0.13\%}}$ \\
$16$ & $7.58$&${\scriptstyle\pm0.48\%}$ & $6.85$&${\scriptstyle\pm0.59\%}$ & $8.06$&${\scriptstyle\pm0.61\%}$ & $\bm{4.77}$&$\bm{{\scriptstyle\pm0.05\%}}$ \\
\bottomrule
\end{tabular}

%% file: tables/table_systems_params.tex
\begin{tabular}{lrrrr}
\toprule
System & HNN & SympNet & FS-HNN & LM-HNN (ours) \\
\midrule
Pendulum & 8,828 & 8,832 & 8,072 & 8,834 \\
Double pendulum & 17,803 & 17,920 & 17,468 & 17,798 \\
FPUT & 26,971 & 26,880 & 26,468 & 26,892 \\
FPUT-Toda & 26,971 & 26,880 & 26,468 & 26,892 \\
FPUT-Morse & 26,971 & 26,880 & 26,468     & 26,892 \\
Fast--slow & 22,041 & 22,016 & 21,488 & 21,957 \\
\bottomrule
\end{tabular}

%% file: tables/table_diag.tex
\begin{tabular}{ll*{3}{r@{}l}}
\toprule
System & Model & \multicolumn{2}{c}{$\sigma_H$} & \multicolumn{2}{c}{$\|\nabla\delta H\|_{\mathrm{rms}}$} & \multicolumn{2}{c}{$k_{\mathrm{eff}}$} \\
\midrule
Pendulum & HNN & $(4.5{\scriptstyle\pm0.8})$&${\scriptstyle\times10^{-4}}$ & $(1.9{\scriptstyle\pm0.4})$&${\scriptstyle\times10^{-3}}$ & $4.1$&${\scriptstyle\pm0.1}$ \\
 & LM-HNN & $\bm{(2.7{\scriptstyle\pm0.9})}$&$\bm{{\scriptstyle\times10^{-5}}}$ & $\bm{(1.1{\scriptstyle\pm0.5})}$&$\bm{{\scriptstyle\times10^{-4}}}$ & $4.2$&${\scriptstyle\pm0.3}$ \\
\midrule
Double pendulum & HNN & $(1.5{\scriptstyle\pm0.2})$&${\scriptstyle\times10^{-2}}$ & $(7.1{\scriptstyle\pm0.7})$&${\scriptstyle\times10^{-2}}$ & $4.8$&${\scriptstyle\pm0.4}$ \\
 & LM-HNN & $\bm{(3.8{\scriptstyle\pm1.5})}$&$\bm{{\scriptstyle\times10^{-4}}}$ & $\bm{(3.4{\scriptstyle\pm1.2})}$&$\bm{{\scriptstyle\times10^{-3}}}$ & $9.1$&${\scriptstyle\pm0.4}$ \\
\midrule
FPUT ($N{=}8$) & HNN & $(1.7{\scriptstyle\pm0.1})$&${\scriptstyle\times10^{-1}}$ & $(3.1{\scriptstyle\pm0.0})$&${\scriptstyle\times10^{-1}}$ & $1.8$&${\scriptstyle\pm0.1}$ \\
 & LM-HNN & $\bm{(4.4{\scriptstyle\pm0.0})}$&$\bm{{\scriptstyle\times10^{-2}}}$ & $\bm{(9.4{\scriptstyle\pm0.7})}$&$\bm{{\scriptstyle\times10^{-2}}}$ & $2.1$&${\scriptstyle\pm0.2}$ \\
\midrule
Perturbed FPUT ($k{=}16$) & HNN & $(2.4{\scriptstyle\pm0.2})$&${\scriptstyle\times10^{-1}}$ & $(1.5{\scriptstyle\pm0.0})$&${\scriptstyle\times10^{0}}$ & $6.3$&${\scriptstyle\pm0.4}$ \\
 & LM-HNN & $\bm{(1.2{\scriptstyle\pm0.1})}$&$\bm{{\scriptstyle\times10^{-1}}}$ & $\bm{(1.4{\scriptstyle\pm0.0})}$&$\bm{{\scriptstyle\times10^{0}}}$ & $12.6$&${\scriptstyle\pm0.8}$ \\
\midrule
FPUT-Toda & HNN & $(2.7{\scriptstyle\pm0.1})$&${\scriptstyle\times10^{-1}}$ & $(4.9{\scriptstyle\pm0.1})$&${\scriptstyle\times10^{-1}}$ & $1.8$&${\scriptstyle\pm0.1}$ \\
 & LM-HNN & $\bm{(3.0{\scriptstyle\pm0.4})}$&$\bm{{\scriptstyle\times10^{-2}}}$ & $\bm{(8.4{\scriptstyle\pm0.4})}$&$\bm{{\scriptstyle\times10^{-2}}}$ & $2.9$&${\scriptstyle\pm0.3}$ \\
\midrule
FPUT-Morse & HNN & $(2.7{\scriptstyle\pm0.3})$&${\scriptstyle\times10^{-1}}$ & $(3.2{\scriptstyle\pm0.1})$&${\scriptstyle\times10^{-1}}$ & $1.2$&${\scriptstyle\pm0.1}$ \\
 & LM-HNN & $\bm{(3.2{\scriptstyle\pm0.5})}$&$\bm{{\scriptstyle\times10^{-2}}}$ & $\bm{(6.1{\scriptstyle\pm0.8})}$&$\bm{{\scriptstyle\times10^{-2}}}$ & $1.9$&${\scriptstyle\pm0.2}$ \\
\midrule[\heavyrulewidth]
System & Model & \multicolumn{2}{c}{\begin{tabular}[b]{@{}c@{}}$\delta H$ power\\beyond $k_{99}$\end{tabular}} & \multicolumn{2}{c}{\begin{tabular}[b]{@{}c@{}}Rollout\\(in range)\end{tabular}} & \multicolumn{2}{c}{\begin{tabular}[b]{@{}c@{}}Rollout\\(above range)\end{tabular}} \\
\midrule
Pendulum & HNN & $12.62$&${\scriptstyle\pm0.63\%}$ & $(2.7{\scriptstyle\pm0.6})$&${\scriptstyle\times10^{-5}}$ & $(1.3{\scriptstyle\pm0.2})$&${\scriptstyle\times10^{0}}$ \\
 & LM-HNN & $87.14$&${\scriptstyle\pm7.02\%}$ & $\bm{(4.1{\scriptstyle\pm2.5})}$&$\bm{{\scriptstyle\times10^{-8}}}$ & $\bm{(2.7{\scriptstyle\pm2.4})}$&$\bm{{\scriptstyle\times10^{-8}}}$ \\
\midrule
Double pendulum & HNN & $2.96$&${\scriptstyle\pm0.33\%}$ & $(4.3{\scriptstyle\pm5.0})$&${\scriptstyle\times10^{-2}}$ & $(1.7{\scriptstyle\pm0.6})$&${\scriptstyle\times10^{0}}$ \\
 & LM-HNN & $87.97$&${\scriptstyle\pm7.35\%}$ & $\bm{(3.2{\scriptstyle\pm3.2})}$&$\bm{{\scriptstyle\times10^{-6}}}$ & $\bm{(8.1{\scriptstyle\pm5.3})}$&$\bm{{\scriptstyle\times10^{-6}}}$ \\
\midrule
FPUT ($N{=}8$) & HNN & $0.85$&${\scriptstyle\pm0.02\%}$ & $(2.0{\scriptstyle\pm0.1})$&${\scriptstyle\times10^{-1}}$ & $(1.8{\scriptstyle\pm0.1})$&${\scriptstyle\times10^{0}}$ \\
 & LM-HNN & $0.78$&${\scriptstyle\pm0.04\%}$ & $\bm{(4.9{\scriptstyle\pm0.6})}$&$\bm{{\scriptstyle\times10^{-3}}}$ & $\bm{(6.5{\scriptstyle\pm1.1})}$&$\bm{{\scriptstyle\times10^{-2}}}$ \\
\midrule
Perturbed FPUT ($k{=}16$) & HNN & $22.11$&${\scriptstyle\pm0.99\%}$ & $(4.2{\scriptstyle\pm0.2})$&${\scriptstyle\times10^{-1}}$ & $(1.6{\scriptstyle\pm0.1})$&${\scriptstyle\times10^{0}}$ \\
 & LM-HNN & $85.04$&${\scriptstyle\pm1.20\%}$ & $\bm{(1.8{\scriptstyle\pm0.0})}$&$\bm{{\scriptstyle\times10^{-1}}}$ & $\bm{(10.0{\scriptstyle\pm1.1})}$&$\bm{{\scriptstyle\times10^{-2}}}$ \\
\midrule
FPUT-Toda & HNN & $0.76$&${\scriptstyle\pm0.01\%}$ & $(7.0{\scriptstyle\pm0.9})$&${\scriptstyle\times10^{-1}}$ & $(1.7{\scriptstyle\pm0.2})$&${\scriptstyle\times10^{0}}$ \\
 & LM-HNN & $0.78$&${\scriptstyle\pm0.11\%}$ & $\bm{(6.7{\scriptstyle\pm4.7})}$&$\bm{{\scriptstyle\times10^{-3}}}$ & $\bm{(1.3{\scriptstyle\pm0.4})}$&$\bm{{\scriptstyle\times10^{-2}}}$ \\
\midrule
FPUT-Morse & HNN & $0.86$&${\scriptstyle\pm0.01\%}$ & $(5.1{\scriptstyle\pm0.5})$&${\scriptstyle\times10^{-1}}$ & $(1.2{\scriptstyle\pm0.1})$&${\scriptstyle\times10^{0}}$ \\
 & LM-HNN & $1.07$&${\scriptstyle\pm0.12\%}$ & $\bm{(1.5{\scriptstyle\pm0.4})}$&$\bm{{\scriptstyle\times10^{-3}}}$ & $\bm{(4.8{\scriptstyle\pm1.5})}$&$\bm{{\scriptstyle\times10^{-3}}}$ \\
\bottomrule
\end{tabular}

%% file: tables/table_fastslow.tex

\begin{tabular}{lr*{4}{r@{}l}}
\toprule
& & \multicolumn{4}{c}{full state} & \multicolumn{4}{c}{slow pair} \\
\cmidrule(lr){3-6}\cmidrule(lr){7-10}
Model & Params & \multicolumn{2}{c}{200 steps} & \multicolumn{2}{c}{1000 steps} & \multicolumn{2}{c}{200 steps} & \multicolumn{2}{c}{1000 steps} \\
\midrule
HNN & 22,041 & $(2.2{\scriptstyle\pm2.6})$&${\scriptstyle\times10^{-3}}$ & $(8.7{\scriptstyle\pm3.7})$&${\scriptstyle\times10^{-3}}$ & $(1.4{\scriptstyle\pm0.2})$&${\scriptstyle\times10^{-1}}$ & $(1.6{\scriptstyle\pm0.2})$&${\scriptstyle\times10^{-1}}$ \\
SympNet & 22,016 & $(1.2{\scriptstyle\pm0.6})$&${\scriptstyle\times10^{-1}}$ & $(1.3{\scriptstyle\pm0.6})$&${\scriptstyle\times10^{0}}$ & $(5.8{\scriptstyle\pm0.4})$&${\scriptstyle\times10^{0}}$ & $(7.4{\scriptstyle\pm0.7})$&${\scriptstyle\times10^{0}}$ \\
FS-HNN & 21,488 & $(1.7{\scriptstyle\pm0.4})$&${\scriptstyle\times10^{-3}}$ & $(2.1{\scriptstyle\pm0.9})$&${\scriptstyle\times10^{-3}}$ & $(1.9{\scriptstyle\pm0.5})$&${\scriptstyle\times10^{-1}}$ & $(3.0{\scriptstyle\pm1.4})$&${\scriptstyle\times10^{-1}}$ \\
\midrule
LM-HNN & 21,957 & $\bm{(2.3{\scriptstyle\pm1.6})}$&$\bm{{\scriptstyle\times10^{-4}}}$ & $\bm{(1.8{\scriptstyle\pm0.5})}$&$\bm{{\scriptstyle\times10^{-3}}}$ & $\bm{(2.8{\scriptstyle\pm0.5})}$&$\bm{{\scriptstyle\times10^{-2}}}$ & $\bm{(4.1{\scriptstyle\pm1.1})}$&$\bm{{\scriptstyle\times10^{-2}}}$ \\
\bottomrule
\end{tabular}